%% file: main.tex
\PassOptionsToPackage{table}{xcolor}
\documentclass[10pt,twocolumn,letterpaper]{article}

\usepackage[pagenumbers]{cvpr} 


\usepackage[table]{xcolor}
\definecolor{cvprblue}{rgb}{0.21,0.49,0.74}
\usepackage[pagebackref,breaklinks,colorlinks,allcolors=cvprblue]{hyperref}
\usepackage{bm}
\usepackage{bbm}
\usepackage{adjustbox}
\usepackage{soul}
\usepackage{multirow}

\def\paperID{601} 
\def\confName{CVPR}
\def\confYear{2025}

\title{SafeguardGS: 3D Gaussian Primitive Pruning While Avoiding Catastrophic Scene Destruction}

\author{Yongjae Lee\\
Arizona State University\\
Tempe, AZ 85281\\
{\tt\small ylee298@asu.edu}
\and
Zhaoliang Zhang\\
Johns Hopkins University\\
Baltimore, MD 21218\\
{\tt\small zzhan288@jh.edu}
\and
Deliang Fan\\
Arizona State University\\
Tempe, AZ 85281\\
{\tt\small dfan@asu.edu}
}

\begin{document}
\maketitle

\begin{abstract}
    3D Gaussian Splatting (3DGS) has made significant strides in novel view synthesis.
    However, its suboptimal densification process results in the excessively large number of Gaussian primitives, which impacts frame-per-second and increases memory usage, making it unsuitable for low-end devices.
    To address this issue, many follow-up studies have proposed various pruning techniques with score functions designed to identify and remove less important primitives.
    Nonetheless, a comprehensive discussion of their effectiveness and implications across all techniques is missing. In this paper, we are the first to categorize 3DGS pruning techniques into two types: \textbf{Scene-level pruning} and \textbf{Pixel-level pruning}, distinguished by their scope for ranking primitives. Our subsequent experiments reveal that, while scene-level pruning leads to disastrous quality drops under extreme decimation of Gaussian primitives, pixel-level pruning not only sustains relatively high rendering quality with minuscule performance degradation but also provides an inherent boundary of pruning, i.e., \textbf{a safeguard of Gaussian pruning}. Building on this observation, we further propose multiple variations of score functions based on the factors of rendering equations and discover that assessing based on \textbf{color similarity with blending weight} is the most effective method for discriminating insignificant primitives. In our experiments, our SafeguardGS with the optimal score function shows the highest PSNR-per-primitive performance under an extreme pruning setting, retaining only about 10\% of the primitives from the original 3DGS scene (i.e., $10\times$ compression ratio). We believe our research provides valuable insights for optimizing 3DGS for future works.
\end{abstract}

\section{Introduction}
\label{sec:intro}

The emergence of Neural Radiance Fields (NeRFs)~\cite{Mildenhall2020NeRF,Barron2021Mipnerf,Barron2022Mipnerf360} has advanced photorealistic 3D reconstruction quality and opened new opportunity for quality enhancement in various fields, such as computer vision~\cite{Xie2021FiGNeRF,Li2023Neuralangelo,Wang2023NeuS2}, robotics~\cite{Yen2021iNeRF,Rosinol2023NeRFSLAM,Zhu2023NICERSLAM}, and extended reality~\cite{Deng2022Fovnerf,Shen2023Envisioning}. The key concept that enables NeRF to achieve photorealistic rendering is encoding 3D scenes with their view-dependent effects into a series of multi-layer perceptrons (MLPs) and synthesizing novel views using alpha-compositing with an appropriate volume sampling strategy (a.k.a. volumetric rendering). However, the deep MLP structure and the high-cost volumetric rendering technique result in substantial computational overhead, hindering prevalent adoption in real-time applications. This has prompted subsequent research to focus on developing specialized data structures~\cite{Yu2021PlenOctrees,Chen2022TensoRF,Fridovich2022Plenoxels,Muller2022Instant}, sophisticated sampling strategies~\cite{Neff2021DONeRF,Kurz2022AdaNeRF,Li2023NerfAcc}, and incorporating deferred rendering techniques~\cite{Hedman2021Baking,Chen2023MobileNeRF} in an effort to improve rendering efficiency.

The recent approach of 3D Gaussian Splatting (3DGS)~\cite{Kerbl20233DGaussian} has demonstrated remarkable success in high-quality 3D scene reconstruction, achieving real-time, high-fidelity scene rendering. Unlike previous NeRF methods~\cite{Mildenhall2020NeRF,Yu2021PlenOctrees,Muller2022Instant}, which implicitly represent 3D scenes using MLPs with dedicated data structures, 3DGS explicitly models scenes with parameterized 3D Gaussian primitives. During rendering, these primitives are efficiently projected and rasterized onto the image plane, a process called Splatting~\cite{Zwicker2001EWA,Zwicker2001Surface}. Their tile-based, differentiable renderer, using an alpha-blending technique, enables both photorealistic rendering and efficient primitive optimization.

\begin{figure}
  \centering
  \includegraphics[width=\linewidth]{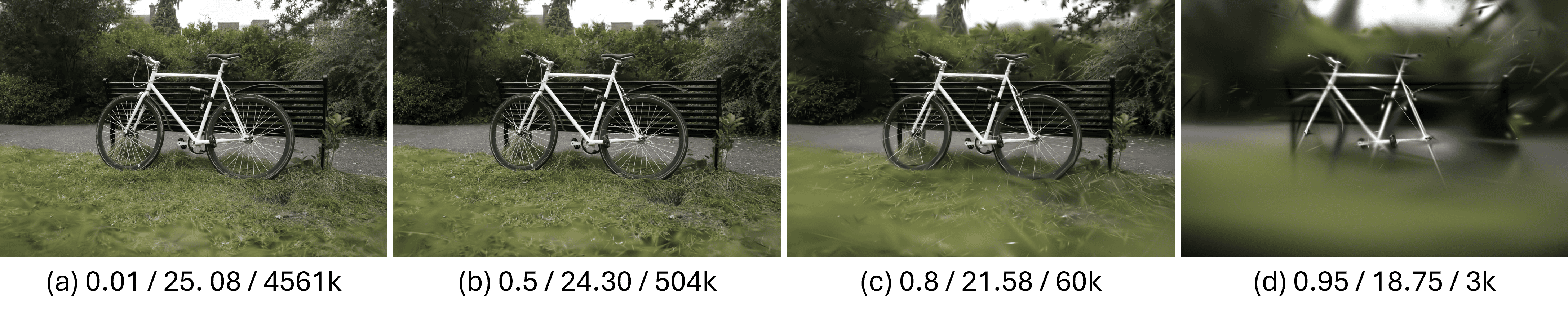}
  \caption{An incautious selection of pruning settings can lead to unexpected scene degradation.
  We reproduce (a)-(d) using the pruning technique from RadSplat~\cite{Niemeyer2024RadSplat}, applied at 20k out of 30k iterations during primitive optimization. The caption of each image indicates importance score threshold (primitives below this are pruned), PSNR, and the number of primitives (Best viewed in zoom).}
  \label{fig:1}
\end{figure}

Despite the diligent efforts of 3DGS to maximize rendering efficiency, it still falls short due to the challenge of primitive booming. During the densification stage, where primitives are cloned and split to rectify erroneously reconstructed regions, the primitives often proliferate excessively, resulting in an overwhelming increase in their number. This explosive growth places heavy demands on computer memory, causing longer rendering latency especially on low-end devices. As noted by~\citet{Fan2023LightGaussian}, most of these primitives contribute little to the rendered output and can be removed without impacting quality. Recognizing this, several recent studies~\cite{Fan2023LightGaussian,Lee2024Compact3D,Fang2024MiniSplatting,Niemeyer2024RadSplat,Liu2024EfficientGS} have proposed various Gaussian primitive pruning techniques that assess the importance of each primitive and eliminate those that are insignificant. However, these techniques have not been thoroughly investigated, leaving the risk of substantial quality degradation when pruning settings are not carefully selected, as illustrated in \cref{fig:1}.

In this paper, we propose SafeguardGS, which inherently imposes a natural limit on the pruning settings, which prevents catastrophic scene destruction, and demonstrates superior performance under extreme pruning conditions compared to existing techniques. Initially, through a thorough examination of existing Gaussian primitive pruning techniques, we identify two major pruning categories: \textit{Scene-level pruning}, which ranks primitives in a global context, and \textit{Pixel-level pruning}, which ranks primitives in an individual ray context.
Interestingly, our preliminary experiment reveals that pixel-level pruning outperforms scene-level pruning in general. Additionally, it suggests that pixel-level pruning inherently indicates a minimum boundary for the number of primitives; surpassing this limit in scene-level pruning causes a catastrophic drop in rendering quality.
Furthermore, we propose multiple score functions that measure each primitive's importance in different aspects and degrees, and benchmark their effectiveness. Our extensive experiments show that color similarity with blending weight can most effectively capture important primitives among the proposed score functions.

We summarize our contributions as:
\begin{itemize}
    \item We classify Gaussian primitive pruning techniques into two categories: scene-level pruning and pixel-level pruning. To the best of our knowledge, this work is the first attempt to categorize these pruning techniques.
    \item We find that pixel-level pruning method structurally has a boundary of pruning, serving as a built-in limit or safeguard. Additionally, we find that pixel-level pruning is more effective at removing insignificant primitives compared to scene-level pruning.
    \item We propose multiple variations of score functions based on the rendering equations and conduct extensive experiments. We discover that color similarity and blending weight together are the most significant factors in discriminating insignificant primitives. In our experiments, the SafeguardGS with the optimal score function shows the highest PSNR-per-primitive performance under an extreme pruning setting, retaining only about 10\% of the primitives from the original 3DGS scene (i.e., $10\times$ compression ratio).
\end{itemize}

\section{Related Work}
\subsection{Efficient neural radiance fields}
In recent decades, numerous methods for reconstructing and representing 3D scenes have attracted significant interest in both computer vision and graphics fields. Traditionally, structure-from-motion (SfM)~\cite{Snavely2006Photo} and multi-view stereo (MVS)~\cite{Goesele2007Multiview} have been utilized to estimate camera parameters from a collection of images and build dense 3D point representations of object shapes in these images. Subsequently, the streamlined software architecture of SfM and MVS, known as COLMAP~\cite{Schoenberger2016Pixelwise, Schoenberger2016Structure}, has become the de facto practice for generating virtual 3D scenes.

Meanwhile, the groundbreaking approach of NeRF~\cite{Mildenhall2020NeRF}, which represents colors and shapes by multi-layer perceptrons (MLPs) encoded features and reconstructs novel view images by synthesizing the features with the volumetric rendering technique, has advanced rendering quality towards photorealism. The success of NeRF has led to its application in various fields, including image processing~\cite{Czerkawski2024Neural}, simultaneous localization and mapping~\cite{Zhu2023NICERSLAM,Rosinol2023NeRFSLAM}, generative modeling~\cite{Niemeyer2021GIRAFFE,Gu2021StyleNeRF,Schwarz2020GRAF}, and human modeling~\cite{Gao2022MPSNeRF,Chen2022GeometryGuided}.

Unsurprisingly, among the many challenges in NeRF, such as reconstructing unbounded 3D scenes~\cite{Zhang2020NeRF++}, data sparsity~\cite{Yu2021pixelNeRF,Wang202IBRNet}, inconsistent data~\cite{Martin2021NeRF}, noisy data~\cite{Ma2022DeblurNeRF}, and temporal data~\cite{Li2021Neural,Gao2021Dynamic}, enhancing the efficiency of both rendering and memory has become a significant concern. A notable contribution to efficient NeRF methods is by~\citet{Sun2022Direct}, who identified the bottleneck in MLPs and proposed a hybrid representation combining feature voxel grids with shallow MLPs. Furthermore,~\citet{Fridovich2022Plenoxels} achieved faster rendering speeds through completely MLP-less feature grids by employing spherical harmonics (SH). Alternatively, other grid-based methods have proposed using more efficient data structures to avoid the high memory strain from dense grids, such as tensor decomposition~\cite{Chen2022TensoRF}, hash tables~\cite{Muller2022Instant}, octrees~\cite{Yu2021PlenOctrees}, and feature quantization~\cite{Shin2024Binary}. Although their endeavors have demonstrated rapid scene convergence and improved image quality compared to traditional MLP-based NeRF methods, grid-based methods are not yet deemed perfect due to limitations in bounded image quality imposed by the selection of data structure for grids and hindered forward-pass speed caused by the inherent multiple sampling approach in the volumetric rendering technique.

\subsection{Compact 3D scene representation with Gaussian primitives}
3D Gaussian Splatting (3DGS)~\cite{Kerbl20233DGaussian} has achieved unprecedented results, establishing a new standard for balancing rendering quality and speed. This remarkable accomplishment is attributed to its MLP-less 3D representation, which utilizes 3D Gaussians to enable real-time, high-quality rendering. Additionally, the explicit 3D scene representation facilitates the use of the typical graphics rendering pipeline, allowing full GPU parallelism.

Despite its state-of-the-art advancement, the densification algorithm of 3DGS~\cite{Kerbl20233DGaussian}, which populates new Gaussian primitives in unreconstructed regions, often leads to millions of primitives occupying vast memory space, posing a major new challenge. Initially, the authors~\cite{Kerbl20233DGaussian} addressed this issue by regularly removing primitives with opacity below a manually pre-selected threshold. However, \citet{Fan2023LightGaussian} demonstrated that this strategy does not effectively reduce the number of primitives. They showed that an additional 60\% pruning, detected through their proposed Gaussian importance evaluation metric, is possible without sacrificing quality. Similarly, concurrent studies have proposed distinct importance metrics to identify unimportant primitives~\cite{Niemeyer2024RadSplat,Fang2024MiniSplatting,Lee2024Compact3D,Liu2024EfficientGS,Zhang2024LP3DGS}; however, the implications of these methods—particularly the key characteristics to consider when designing such metrics—remain unveiled.

\section{Gaussian Primitive Pruning}
\subsection{Background}\label{sec:background}
\paragraph{3D Gaussian Splatting.}
3DGS~\cite{Kerbl20233DGaussian} represents a 3D scene using a set of \(N\) 3D Gaussians as geometric primitives. Gaussians \(\{\mathbf{G}_i \mid i=1,2,3,\ldots,N\}\) are parameterized by center position (mean) \(\bm{\mu_i} \in \mathbb{R}^3\), scale \(\mathbf{s}_i \in \mathbb{R}^3\), rotation \(\mathbf{r}_i \in \mathbb{R}^4\), opacity \(\mathbf{o}_i \in \mathbb{R}\), and \(k\) spherical harmonics (SH) coefficients \(\mathbf{f}_i \in \mathbb{R}^k\). In practice, the opacity value is activated by a sigmoid function, i.e., \(\bm{\sigma}_i = \sigma(\mathbf{o}_i)\), to bound the range to \([0,1]\), and the degree of the SH is set to three. Eventually, the total number of parameters of a Gaussian is 59. 

In the initialization of the Gaussian primitives, 3DGS uses an SfM-generated sparse point cloud~\cite{Snavely2006Photo} if available, or alternatively, it initializes a fixed number of Gaussians with positions and colors randomly drawn from a uniform distribution. During the first half of training iterations, the densification process is employed to augment the number of primitives, addressing improperly represented geometries (e.g., under-reconstruction and over-reconstruction) and inadequately emerged floaters. After the densification stage, the number of primitives remains constant, and only the remaining primitives are further optimized.

3DGS employs the splatting technique~\cite{Zwicker2001EWA,Zwicker2001Surface} to render images. Each 3D Gaussian with its center \(\bm{\mu}\) is defined as \(G(\mathbf{x}) = \exp(-\frac{1}{2} \mathbf{x}^T\mathbf{\Sigma}^{-1} \mathbf{x})\), where the covariance matrix \(\mathbf{\Sigma}\) in world coordinates is given by \(\mathbf{RS}\mathbf{S}^{T}\mathbf{R}^{T}\). Here, \(\mathbf{R}\) and \(\mathbf{S}\) are the rotation and scale matrices derived from their respective vector forms. Each 3D Gaussian is then projected onto the image plane as a 2D splat \(\mathcal{G}\) by omitting the last row and column of the transformed covariance in camera coordinates: \(\mathbf{\Sigma'} = \mathbf{JW\Sigma}\mathbf{W}^T\mathbf{J}^T\), where \(\mathbf{W}\) is the view transformation matrix, and \(\mathbf{J}\) is the Jacobian of the affine-approximated projective transformation. Finally, the color \(\mathbf{c}_r\) of each pixel ray \(r \in \mathcal{R}\) is determined by progressively accumulating the colors \(c_i\) of the projected \(M\) splats on the pixel, weighted by \(w_i\), from the nearest to the farthest splats from the camera. Formally,

\begin{equation}
    \mathbf{c}_r = \sum_{i \in M} c_i w_i, \label{eq:rendering_function}
\end{equation}
\begin{equation*}
    \text{with } w_i = \alpha_i T_i, \quad \alpha_i = \bm{\sigma}_i \mathcal{G}_i \text{, \quad and } T_i = \prod_{j=1}^{i-1}(1-\alpha_j).
\end{equation*}

The view-dependent appearance \(c_i\) of Gaussian primitive \(\mathbf{G}_i\) is computed from its SH coefficients \(\mathbf{f}_i\).

\begin{figure*}[!ht]
  \centering
  \includegraphics[width=\linewidth]{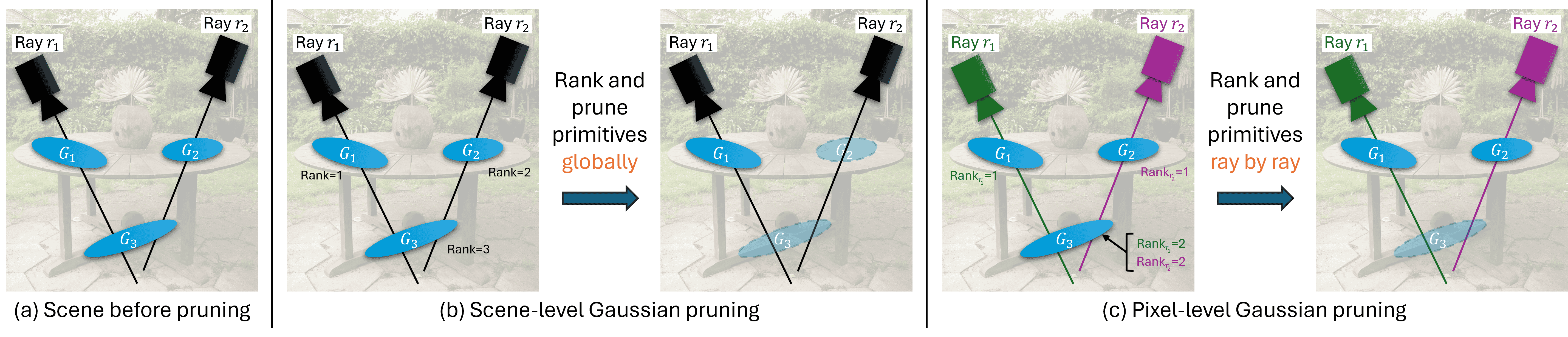}
  \caption{Given a set of Gaussian primitives \(\mathbf{G}_i\) (a), scene-level pruning (b) evaluates the importance of each primitive across all rays and orders the primitives globally. Because this pruning is applied at the scene level, some rays may end up rendering nothing afterward. In contrast, pixel-level pruning (c) evaluates each primitive's contribution and ranks the primitives accordingly independently for each ray, ensuring that primitives are preserved in a way that completely prevents rays from missing intersections with primitives.}
  \label{fig:2}
\end{figure*}

\paragraph{Gaussian primitive pruning.}
The required model size to represent a 3D scene with Gaussian primitives increases proportionally with the number of primitives. To control the number of primitives, 3DGS introduces the densification process. However, its naive approach of duplicating and removing primitives seems suboptimal and tends to proliferate millions of primitives, as discussed by the authors~\cite{Kerbl20233DGaussian}. For example, a scene represented by one million primitives with full-precision parameters results in a memory pressure of approximately 236MB (= 59\(\times\)4Byte\(\times\)1M). Moreover, since the differentiable renderer orders the primitives and accesses them from the nearest to the farthest on a pixel-by-pixel basis, an excessive number of primitives critically hampers rendering speed as well as burdens rendering memory.

Recent works recognize this problem and propose methods to prune primitives based on specific criteria. The common approach involves defining a metric to quantify each primitive's importance and generating a mask for retaining primitives that meet a specific \textit{pruning setting} (such as pruning ratio, score threshold, or top-k rank) based on the evaluated importance. For instance, LightGaussian~\cite{Fan2023LightGaussian} defines a global significance score depending on a primitive's opacity \(\mathbf{o}_i\) and adaptively normalized volume \(\gamma_i = (\min(\max(\frac{\prod \mathbf{s}_i}{\prod \mathbf{s}_{max90}},0),1)^{0.1}\), across all training views, and prunes primitives whose importance falls below a pre-selected pruning ratio. Similarly, Mini-Splatting~\cite{Fang2024MiniSplatting} defines an importance score that primarily measures a primitive's contribution (i.e., \(w_i\) in \cref{eq:rendering_function}) to the rendered pixel colors. The number of primitives is controlled by a predefined pruning ratio, but unlike LightGaussian, Mini-Splatting uses importance scores as the sampling probability for each primitive, and they are stochastically sampled to better maintain overall geometry with the pruning ratio. However, accumulating the importance of each primitive to every pixel may exaggerate importance when a primitive has a high hit count. Meanwhile, RadSplat~\cite{Niemeyer2024RadSplat} proposes using the max operator instead of summation and identifies low priority primitives when their maximum contribution across rays falls below a predefined score threshold. Similarly, EfficientGS~\cite{Liu2024EfficientGS} employs the same idea of measuring primitive's contribution using blending weight \(w_i\) but with a different pruning setting, independently determining the top-k dominant primitives with significant contributions for each pixel and pruning those ranked lower than the predefined k-th rank. The score functions used in these studies are summarized as follows:
\begin{align*}
    Score_{LG}(\mathbf{G}_i, \mathcal{R}) &= \sum_{r \in \mathcal{R}} \mathbbm{1}(\mathbf{G}_i, r) \cdot \bm{\sigma}_i \cdot \gamma_i, \quad &\text{\cite{Fan2023LightGaussian}} \\
    Score_{MS}(\mathbf{G}_i, \mathcal{R}) &= \sum_{r \in \mathcal{R}} \mathbbm{1}(\mathbf{G}_i, r) \cdot w_i, \quad &\text{\cite{Fang2024MiniSplatting}} \\
    Score_{RS}(\mathbf{G}_i, \mathcal{R})& = \max_{r \in \mathcal{R}} \mathbbm{1}(\mathbf{G}_i, r) \cdot w_i, \quad &\text{\cite{Niemeyer2024RadSplat}} \\
    Score_{EG}(\mathbf{G}_i, r) &= w_i. \quad &\text{\cite{Liu2024EfficientGS}}
\end{align*}
Here, \(\mathbbm{1}(\mathbf{G}_i, r)\) represents an indicator function that specifies whether the Gaussian primitive \(\mathbf{G}_i\) intersects with the ray \(r\) or not. \(Score(\mathbf{G}_i, \mathcal{R})\) evaluates the contribution of the primitive across all rays \(\mathcal{R}\), while \(Score(\mathbf{G}_i, r)\) assesses its contribution with respect to a specific ray \(r\).

\subsection{Categorization of Gaussian primitive pruning} \label{sec:categorization}
We reviewed the contemporary Gaussian primitive pruning efforts in~\cref{sec:background}. The aforementioned pruning techniques utilize distinct score functions and pruning settings. Essentially, they can be categorized into two types depending on whether the score evaluation considers all rays collectively or each ray independently.

\paragraph{Scene-level pruning.}
One type of pruning involves evaluating the Gaussian primitives \(\mathbf{G}_i\) at a global level~\cite{Fan2023LightGaussian, Fang2024MiniSplatting,Niemeyer2024RadSplat}. As the term `scene-level' implies, each primitive's contribution is evaluated across all pixels, either by aggregating or maximizing its impact on the entire scene. The contribution is measured in various ways, with activated opacity \(\bm{\sigma}_i\) and blending weight \(w_i\) being commonly used. Once the scores are computed, primitives that meet a criterion---such as a certain percentage of top-scored primitives, those with scores above a threshold, or the top k-ranked primitives---are retained (\cref{fig:2}~(b)).

\paragraph{Pixel-level pruning.}
Another type of pruning involves comparing the importance of primitives within the scope of each pixel ray~\cite{Liu2024EfficientGS}. In this method, a group \(M\) of primitives intersecting a ray \(r\) is first identified. Each primitive's contribution to the ray is measured based on its parameters, similar to scene-level pruning. Then, those that meet a defined criterion are marked for retention. This pixel-level masking is repeated for every ray \(\mathcal{R}\). The resulting masks are typically merged using an `or' operation, and primitives are removed based on this final mask (\cref{fig:2} (c)).


\paragraph{Scene-level vs. pixel-level.}
The main difference between these two pruning types lies in their scope, which gives pixel-level pruning some unique advantages. Scene-level pruning ranks primitives based on their global importance, which can lead to potential over- or underestimation of importance depending on the number of rays that intersect the primitive. For example, consider a primitive that makes a small, uniform contribution across all rays; its absence would not impact rendering quality. In contrast, consider another primitive that has a high contribution but intersects only a single ray. The Mini-Splatting score function (\(Score_{MS}\)) might assign a higher score to the first primitive due to its broader exposure, prioritizing it over the second, which could then be pruned. In pixel-level pruning, however, importance is evaluated per ray, allowing the second primitive to retain its significance to the intersecting ray and increasing the likelihood that the first, less critical primitive will be pruned.

Another advantage of pixel-level pruning is that it guarantees at least one primitive per ray, ensuring every ray renders some content, rather than blank. Intuitively, an extreme pixel-level pruning setting could retain only the highest-contributing primitive for each ray. This provides a natural boundary for pruning since removing any more would result in an empty scene, which is nonsensical. Scene-level pruning, by contrast, lacks such safeguard and carries the risk of leaving some rays with no intersecting primitives if pruning settings are chosen improperly. For instance, if only the top single contributing primitive is kept across the entire scene, many rays may render nothing or only display the same primitive. To avoid this issue, empirically sweeping pruning parameters is essential to find the proper settings with evitable large computing overhead.

\begin{figure}
  \centering
  \includegraphics[width=1\linewidth]{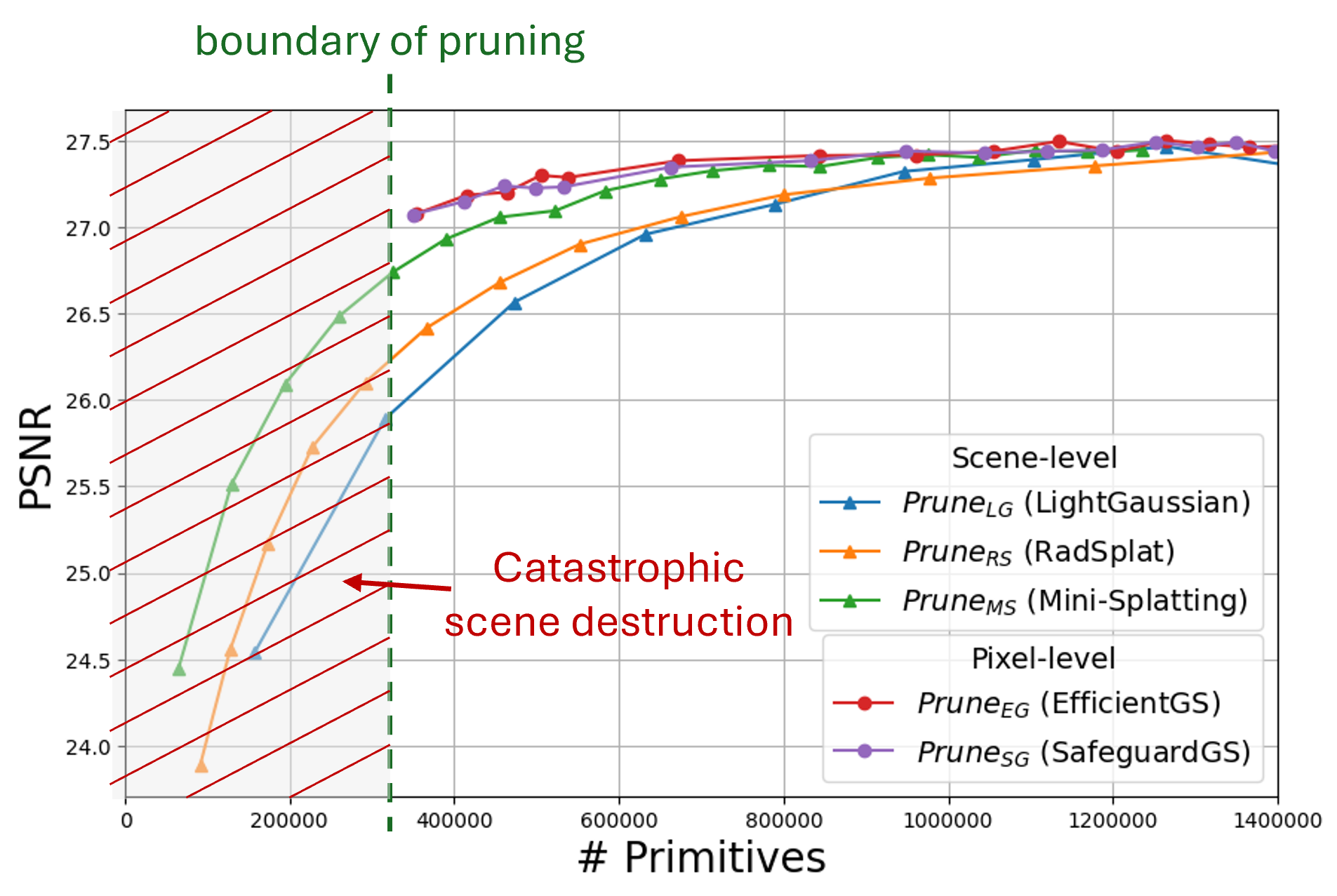}
  \caption{Averaged PSNR over nine scenes from MipNeRF360~\cite{Barron2022Mipnerf360}. We sweep pruning parameters for each method. Unlike scene-level pruning, pixel-level pruning maintains the number of primitives above a certain boundary, preventing drastic scene degradation while preserving high quality.}
  \label{fig:3}
\end{figure}

To explore the implications of the two types of pruning, we train 3DGS for 30k iterations and apply pruning techniques~\cite{Fan2023LightGaussian,Fang2024MiniSplatting,Niemeyer2024RadSplat,Liu2024EfficientGS} at 20k iteration, each with sweeping pruning parameters.
\cref{fig:3} shows that the novel view rendering quality according to the number of primitives. We observe that the PSNR value of both pruning types begins to drop at around 0.6 million primitives. However, their slopes differ. While scene-level pruning techniques indiscriminately prune until nothing remained, pixel-level pruning techniques avoid over-decimation of the primitives, maintaining at least around 0.35 million primitives. Moreover, pixel-level pruning always shows better quality compared to scene-level pruning. Apparently, \ul{pixel-level pruning prevents inaccurate estimation of importance as well as the catastrophic scene destruction.}

\begin{figure}
  \centering
  \includegraphics[width=\linewidth]{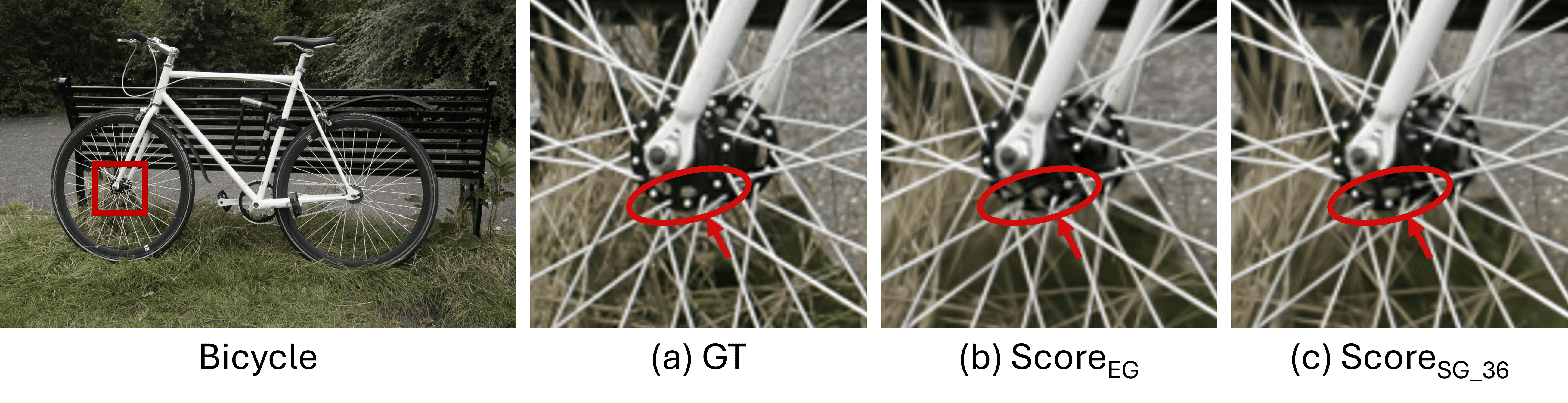}
  \caption{The score function of EfficientGS (\(Score_{EG}\)) lacks consideration of the colors of Gaussian primitives, resulting in the loss of high-frequency information during pruning, as indicated by the white holes on the wheel axle in (b). Our proposed score function (\(Score_{SG\_36}\)), which is sensitive to the color similarity between the primitive color and the ground truth pixel value, helps preserve the details in (c).}
  \label{fig:4}
\end{figure}

\subsection{Importance score functions}\label{sec:score_functions}
Both scene-level pruning and pixel-level pruning techniques require defining a score function assessing the importance of a primitive, as discussed in \cref{sec:background}. Furthermore, we have demonstrated in \cref{sec:categorization} that under aggressive pruning settings, pixel-level pruning can reasonably preserve relatively high quality (EfficientGS with a top-1 criterion in \cref{fig:3} achieves a PSNR of 27.07 dB with 353k primitives). However, this is still lower compared to 3DGS (PSNR of 27.43 dB). \cref{fig:4}~(a) and (b) compare the rendered image of fully trained 3DGS and the rendered image immediately after pruning by using EfficientGS pruning with a top-1 criterion. This juxtaposition suggests that the score function of EfficientGS is insufficiently sensitive to the color difference between that of primitive and that of the ground truth, resulting in the loss of high-frequency areas when the primitives are pruned too aggressively.

Revisiting the rendering equation, ~\cref{eq:rendering_function}, we notice that the following Gaussian parameters and intermediate values contribute to the rendered color of a pixel: \(c_i, \bm{\sigma}_i, \alpha_i, \mathcal{G}_i,\) and \(T_i\). Based on these factors, we define various score functions by combining them with appropriate activation of color similarity and pixel closeness, listed as in \cref{table:6} (we list part of the proposed score functions here, and please see supplementary material for the full list).

\begin{table}[]
\centering
\caption{Selected list of proposed score functions}
\label{table:6}
\resizebox{\columnwidth}{!}{%
\begin{tabular}{@{}l@{}}
\toprule
\(Score_{SG\_3}(\mathbf{G}_i, r) = \bm{\sigma}_i * T_i\)                           \\
\(Score_{SG\_4}(\mathbf{G}_i, r) = \alpha_i * T_i\) (equivalent to \(Score_{EG}\)) \\
\(Score_{SG\_8}(\mathbf{G}_i, r) = PC(X_r, X_i) *  \bm{\sigma}_i * T_i\)            \\
\(Score_{SG\_9}(\mathbf{G}_i, r) = PC(X_r, X_i) * \alpha_i * T_i\)                   \\
\(Score_{SG\_11}(\mathbf{G}_i, r) = \alpha_i + T_i\)        \\
\(Score_{SG\_36}(\mathbf{G}_i, r) = CS_{1}(\mathbf{c}_{r}, c_i) * \alpha_i * T_i\)  \\
\(Score_{SG\_39}(\mathbf{G}_i, r) = CS_{1}(\mathbf{c}_{r}, c_i) * PC(X_r, X_i) * \alpha_i\) \\
\(Score_{SG\_40}(\mathbf{G}_i, r) = CS_{2}(\mathbf{c}_{r}, c_i) * PC(X_r, X_i) * \bm{\sigma} * T_i\) \\
\(Score_{SG\_52}(\mathbf{G}_i, r) = CS_{2}(\mathbf{c}_{r}, c_i) * \alpha_i * T_i\) \\
\(Score_{SG\_57}(\mathbf{G}_i, r) = CS_{2}(\mathbf{c}_{r}, c_i) * PC(X_r, X_i) * \alpha_i + T_i\) \\
\(Score_{SG\_68}(\mathbf{G}_i, r) = CS_{1}(\mathbf{c}_{r}, c_i) + \alpha_i * T_i\)        \\
\(Score_{SG\_75}(\mathbf{G}_i, r) = CS_{1}(\mathbf{c}_{r}, c_i) + \alpha_i + T_i\)        \\
\(Score_{SG\_84}(\mathbf{G}_i, r) = CS_{2}(\mathbf{c}_{r}, c_i) + \alpha_i * T_i\)        \\
\(Score_{SG\_91}(\mathbf{G}_i, r) = CS_{2}(\mathbf{c}_{r}, c_i) + \alpha_i + T_i\)        \\ \bottomrule
\end{tabular}%
}
\parbox{\columnwidth}{\vspace{2pt}\footnotesize{* Refer to \cref{sec:background} and \cref{sec:score_functions} for the meaning of the letters (\(\bm{\sigma}_i\), \(\alpha_i\), and \(T_i\)) and functions (\(PC\), \(CS_1\), and \(CS_2\)).}}
\end{table}

Here, \(PC(\cdot,\cdot)\), \(CD_{1}(\cdot,\cdot)\), and \(CD_{2}(\cdot,\cdot)\) are defined functions for pixel closeness and color similarity, specifically:
\begin{align}
    PC(X_r, X_i) &= \exp(-||X_r - X_i||_2), \notag \\
    CS_{1}(\mathbf{c}_r, c_i) &= (1 - mean(||\mathbf{c}_r - c_i||_1)), \label{eq:2} \\
    CS_{2}(\mathbf{c}_r, c_i) &= \exp (- mean(||\mathbf{c}_r - c_i||_1)), \notag
\end{align}
where \(X_r\) and \(\mathbf{c}_r\) are the pixel coordinates and color of the ray \(r\), respectively, and \(X_i\) is the center position of the projected splat \(\mathcal{G}_i\).

Through extensive experiments (\cref{sec:exp_imp_score_function}), we discover that \(Score_{SG\_36}\), which consists of color similarity and blending weight, preserves better high-frequency areas compared to \(Score_{EG}\), as depicted in \cref{fig:4}~(b) and (c).

\begin{table*}[]
\centering
\caption{PSNR-per-primitive comparison on MipNeRF360~\cite{Barron2022Mipnerf360}. The darker the pink color, the better the effectiveness of pruning.}
\label{table:1}
\resizebox{\textwidth}{!}{%
\begin{tabular}{@{}clrrrrrrrrrrrrrrr@{}}
\toprule
\multicolumn{1}{l}{}         &                           & \multicolumn{15}{c}{Top-k}                                                                                                                                                                                                                                                                                                                                                       \\ \cmidrule(l){3-17}
\multicolumn{1}{l}{}         &                           & \multicolumn{1}{c}{1} & \multicolumn{1}{c}{2} & \multicolumn{1}{c}{3} & \multicolumn{1}{c}{4} & \multicolumn{1}{c}{5} & \multicolumn{1}{c}{10} & \multicolumn{1}{c}{20} & \multicolumn{1}{c}{30} & \multicolumn{1}{c}{40} & \multicolumn{1}{c}{50} & \multicolumn{1}{c}{60} & \multicolumn{1}{c}{70} & \multicolumn{1}{c}{80} & \multicolumn{1}{c}{90} & \multicolumn{1}{c}{100} \\ \midrule
\multirow{3}{*}{\(Score_{SG\_4}\)} & PSNR\(\uparrow\)          & 27.074                & 27.184                & 27.201                & 27.298                & 27.286                & 27.383                 & 27.413                 & 27.416                 & 27.439                 & 27.495                 & 27.437                 & 27.500                 & 27.479                 & 27.461                 & 27.469                  \\
                             & \# Prim.\(\downarrow\) (k)  & 353.134               & 415.036               & 463.817               & 505.024               & 538.258               & 671.651                & 843.440                & 960.525                & 1055.747               & 1133.836               & 1204.023               & 1264.763               & 1316.377               & 1365.480               & 1413.110                \\
                             & PSNR/Prim.\(\uparrow\) & \cellcolor{pink!50}7.67E-5              & \cellcolor{pink!50}6.55E-5              & \cellcolor{pink!50}5.86E-5              & \cellcolor{pink!50}5.41E-5              & \cellcolor{pink!50}5.07E-5              & \cellcolor{pink!50}4.08E-5               & \cellcolor{pink!50}3.25E-5               & \cellcolor{pink!50}2.85E-5               & \cellcolor{pink!50}2.60E-5               & \cellcolor{pink!50}2.42E-5               & \cellcolor{pink!50}2.28E-5               & \cellcolor{pink!50}2.17E-5               & \cellcolor{pink!50}2.09E-5               & \cellcolor{pink!50}2.01E-5               & \cellcolor{pink!50}1.94E-5                \\ \midrule
\multirow{3}{*}{\(Score_{SG\_36}\)} & PSNR\(\uparrow\)          & 27.073                & 27.148                & 27.238                & 27.225                & 27.232                & 27.346                 & 27.384                 & 27.439                 & 27.427                 & 27.438                 & 27.445                 & 27.492                 & 27.462                 & 27.492                 & 27.438                  \\
                             & \# Prim.\(\downarrow\) (k)  & 350.294               & 412.105               & 459.782               & 498.216               & 533.089               & 662.526                & 832.465                & 948.367                & 1043.723               & 1120.808               & 1186.437               & 1251.024               & 1303.052               & 1349.590               & 1396.247                \\
                             & PSNR/Prim.\(\uparrow\) & \cellcolor{pink}7.73E-5              & \cellcolor{pink}6.59E-5              & \cellcolor{pink}5.92E-5              & \cellcolor{pink}5.46E-5              & \cellcolor{pink}5.11E-5              & \cellcolor{pink}4.13E-5               & \cellcolor{pink}3.29E-5               & \cellcolor{pink}2.89E-5               &\cellcolor{pink} 2.63E-5               & \cellcolor{pink}2.45E-5               & \cellcolor{pink}2.31E-5               & \cellcolor{pink}2.20E-5               & \cellcolor{pink}2.11E-5               & \cellcolor{pink}2.04E-5               & \cellcolor{pink}1.97E-5                \\ \bottomrule
\end{tabular}%
}
\end{table*}

\subsection{SafeguardGS}
We have discussed two key advantages of pixel-level pruning and found that color similarity with blending weight is most effective for scoring primitives. Building on these insights, we propose an optimal pruning technique for 3DGS, called SafeguardGS. SafeguardGS quantifies the significance of each primitive using \(Score_{SG\_36}\) and prunes according to a top-1 criterion pixel by pixel.



\section{Experiment}
\subsection{Setup}
\paragraph{Implementation details.}
Thanks to the open-source code of 3DGS~\cite{Kerbl20233DGaussian}, we readily built our experimental platform. We implemented the primitive pruning algorithm on top of 3DGS's Python code and extended the CUDA-based differentiable renderer to include the various score functions.
We trained each technique for 30k iterations on an NVIDIA A100 GPU. During training, we applied pruning once at 20k iterations for a fair comparison. For other hyperparameters (e.g., learning rates, SH degree scheduling, and densification process), we followed the 3DGS configuration to maintain consistency across all pruning techniques and datasets.

\paragraph{Datasets and Metrics.}
For the comprehensive comparison of pruning techniques and the effectiveness of importance score functions, we train each pruning technique on various datasets including Synthetic NeRF~\cite{Mildenhall2020NeRF}, MipNeRF360~\cite{Barron2022Mipnerf360}, Tanks\&Temples~\cite{Knapitsch2017Tanks}, and Deep Blending~\cite{Hedman2018Deep}, and we evaluate the rendering quality by using PSNR, SSIM, and LPIPS. Additionally, we report the number of primitives in thousands, the ratio of pruned primitives, and PSNR-per-primitive.

We follow the common approaches for using the datasets. For the Synthetic NeRF dataset, comprising eight small, bounded virtual scenes, we allocate one hundred 800x800 images for training and two hundred 800x800 images for testing. For the unbounded, real-world scenes in the MipNeRF360 dataset (consisting of four indoor scenes and five outdoor scenes), the Tanks\&Temples dataset (containing two outdoor scenes), and the Deep Blending dataset (containing two indoor scenes), we use every 8th image for testing and the rest for training at their original resolutions.

\paragraph{Baselines for pruning technique comparison.}
As our motivation to comprehend the implications of different pruning techniques, we have imported the only the pruning implementations (i.e., pruning setting and score function) from the prior works. The pruning technique notations are \(Prune_{LG}\), \(Prune_{MS}\), \(Prune_{RS}\), \(Prune_{EG}\), and \(Prune_{SG}\), originating respectively from LightGaussian~\cite{Fan2023LightGaussian}, Mini-Splatting~\cite{Fang2024MiniSplatting}, RadSplat~\cite{Niemeyer2024RadSplat}, and EfficientGS~\cite{Liu2024EfficientGS}, and our proposed method.
Note that, as of this paper's submission, only LightGaussian and Mini-Splatting have open-sourced their codes, so we carefully implemented the pruning techniques for RadSplat and EfficientGS based on the descriptions in their respective papers.

\subsection{Study of effective score function}\label{sec:exp_imp_score_function}
Firstly, we conducted experiments to identify the most effective score function among all proposed ones. For each function proposed in~\cref{table:6}, we swept the top-k criterion from 1 to 100, recording the corresponding reconstruction quality. As expected, higher top-k values allow more primitives to remain, resulting in a larger number of primitives after pruning and thus higher quality. However, the effectiveness of each function varies; some retain primitives that contribute more significantly to reconstruction quality, meaning that a comparable number of primitives yields higher reconstruction fidelity.

\begin{figure}
  \centering
  \includegraphics[width=\linewidth]{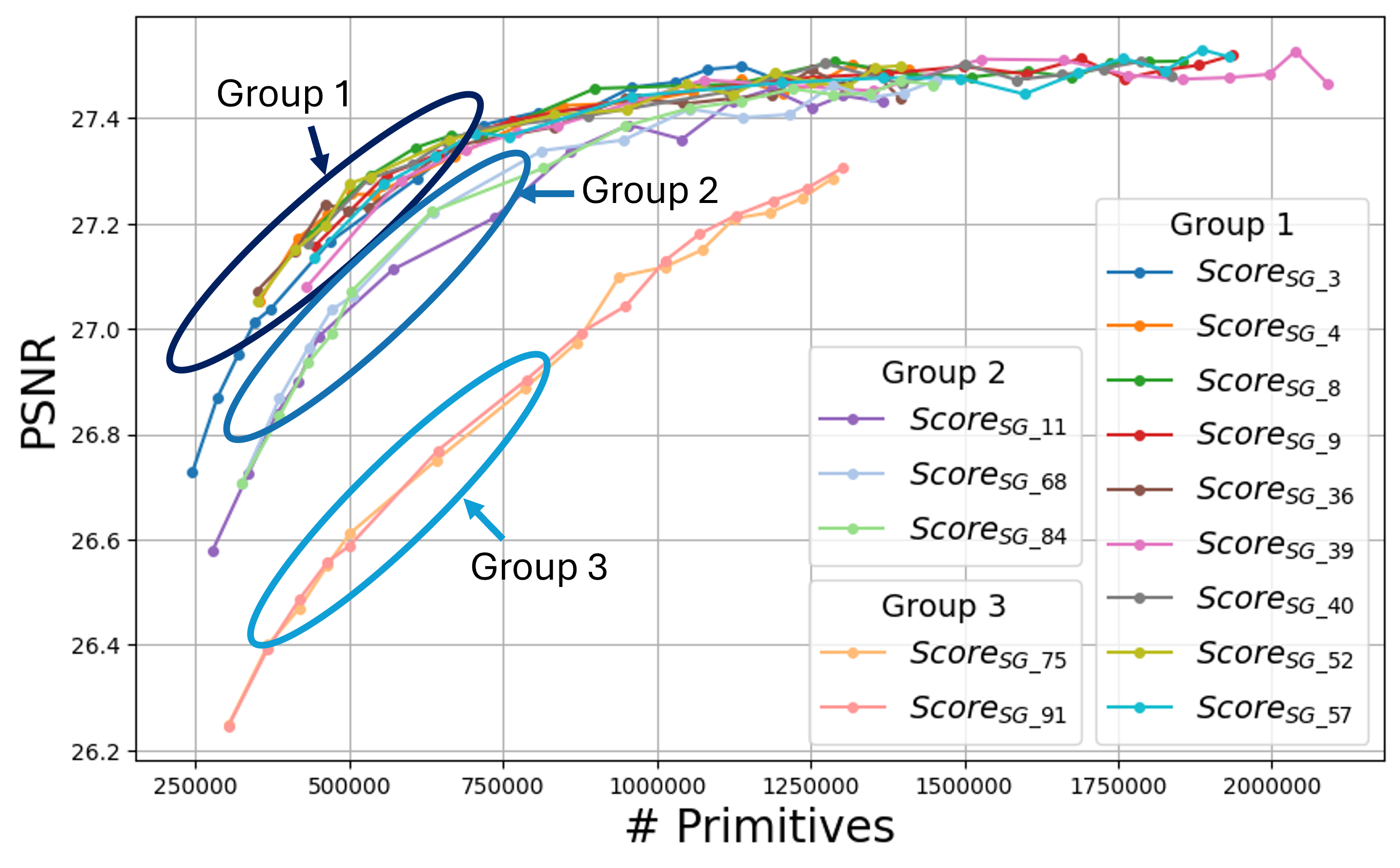}
  \caption{We perform a top-k parameter sweep across a group of proposed score functions, as listed in \cref{table:1}.
  Results are averaged over nine scenes from MipNeRF360~\cite{Barron2022Mipnerf360}.}
  \label{fig:5}
\end{figure}

In our analysis, we found that the blending weight \(w_i\) effectively prioritizes high-quality primitives, positioning curves of \(Scor_{SG\_4}\) and \(Scor_{SG\_36}\) at the top in~\cref{fig:5}. For further results from additional metrics and datasets, please refer to supplementary material.

\begin{figure}
  \centering
  \includegraphics[width=\linewidth]{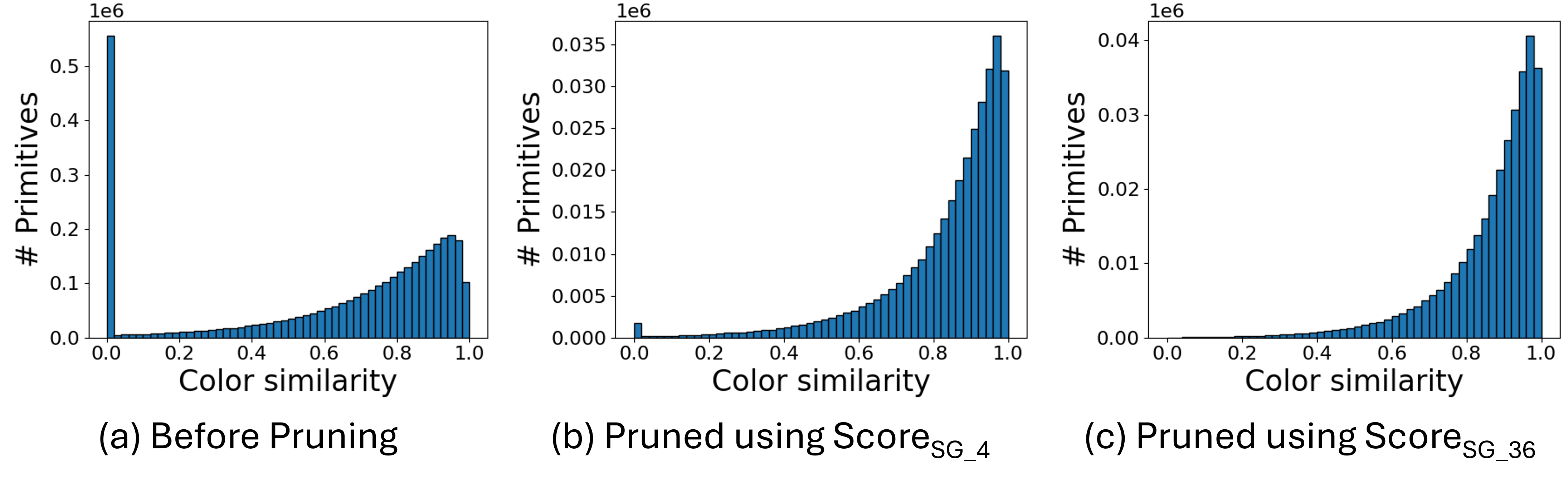}
  \caption{Histograms of color similarity are shown (a) before pruning, (b) after pruning with \(Score_{SG\_4}\), and (c) with \(Score_{SG\_{36}}\). Note that while \(Score_{SG\_4}\) leaves some invalidated primitives which are having nearly zero color similarity, \(Score_{SG\_{36}}\) perfectly prunes them. This graph is collected from the flowers scene in MipNeRF360~\cite{Barron2022Mipnerf360}.}
  \label{fig:6}
\end{figure}

To better understand the impact of color similarity, we analyzed the distribution of primitives' color similarity. \cref{fig:6} demonstrates that lots of primitives with low color similarity are removed by \(Scor_{SG\_4}\), while \(Scor_{SG\_36}\) completely eliminates all primitives with low color similarity. For other scenes, refer to supplementary material.

\cref{table:1} displays PSNR-per-primitive across different top-k values, revealing that \(Scor_{SG\_36}\) effectively prunes thousands of primitives while maintaining nearly the same reconstruction quality as \(Scor_{SG\_4}\). We observed that the PSNR-per-primitive of \(Scor_{SG\_36}\) consistently exceeds that of other functions, concluding that \(Scor_{SG\_36}\) is the optimal choice for effective pruning.

\begin{table}[]
\centering
\caption{Result on Synthetic NeRF~\cite{Mildenhall2020NeRF}. The darker the pink color, the better the effectiveness of pruning.}
\label{table:2}
\resizebox{\columnwidth}{!}{%
\begin{tabular}{@{}lcccccc@{}}
\toprule
Prune          & PSNR\(\uparrow\) & SSIM\(\uparrow\) & LPIPS\(\downarrow\) & \begin{tabular}[c]{@{}c@{}}\# Prim.\(\downarrow\)\\ (k)\end{tabular} & Comp. ratio\(\uparrow\) & PSNR/Prim.\(\uparrow\) \\ \midrule
\(Prune_{LG}\)  & 31.416           & 0.9567           & 0.0490              & 43.4                                                                 & 85.0\%                  & 7.24E-4                \\
\(Prune_{MS}\) & 33.086           & 0.9666           & 0.0360              & 37.1                                                                 & 87.2\%                  & \cellcolor{pink!20}8.93E-4                \\
\(Prune_{RS}\)       & 32.787           & 0.9643           & 0.0411              & 38.4                                                                 & 86.7\%                  & 8.54E-4                \\
\(Prune_{EG}\)    & 32.910           & 0.9658           & 0.0369              & 33.8                                                                 & 88.3\%                  & \cellcolor{pink}9.74E-4                \\
\(Prune_{SG}\)    & 32.876           & 0.9657           & 0.0371              & 33.8                                                                 & 88.3\%                  & \cellcolor{pink!50}9.73E-4                \\ \bottomrule
\end{tabular}%
}
\end{table}

\begin{table}[]
\centering
\caption{Result on MipNeRF360~\cite{Barron2022Mipnerf360}. The darker the pink color, the better the effectiveness of pruning.}
\label{table:3}
\resizebox{\columnwidth}{!}{%
\begin{tabular}{@{}lcccccc@{}}
\toprule
Prune          & PSNR\(\uparrow\) & SSIM\(\uparrow\) & LPIPS\(\downarrow\) & \begin{tabular}[c]{@{}c@{}}\# Prim.\(\downarrow\)\\ (k)\end{tabular} & Comp. ratio\(\uparrow\) & PSNR/Prim.\(\uparrow\) \\ \midrule
\(Prune_{LG}\)  & 26.565           & 0.7794           & 0.2767              & 473.1                                                                & 85.0\%                  & 5.62E-5                \\
\(Prune_{MS}\) & 26.929           & 0.7956           & 0.2612              & 389.4                                                                & 87.7\%                  & 6.92E-5                \\
\(Prune_{RS}\)       & 26.415           & 0.7840           & 0.2833              & 366.0                                                                & 88.4\%                  & \cellcolor{pink!20}7.22E-5                \\
\(Prune_{EG}\)    & 27.074           & 0.7889           & 0.2674              & 353.1                                                                & 88.8\%                  & \cellcolor{pink!50}7.67E-5                \\
\(Prune_{SG}\)    & 27.073           & 0.7881           & 0.2688              & 350.3                                                                & 88.9\%                  & \cellcolor{pink}7.73E-5                \\ \bottomrule
\end{tabular}%
}
\end{table}

\begin{table}[]
\centering
\caption{Result on TanksAndTemples~\cite{Knapitsch2017Tanks}. The darker the pink color, the better the effectiveness of pruning.}
\label{table:4}
\resizebox{\columnwidth}{!}{%
\begin{tabular}{@{}lcccccc@{}}
\toprule
Prune          & PSNR\(\uparrow\) & SSIM\(\uparrow\) & LPIPS\(\downarrow\) & \begin{tabular}[c]{@{}c@{}}\# Prim.\(\downarrow\)\\ (k)\end{tabular} & Comp. ratio\(\uparrow\) & PSNR/Prim.\(\uparrow\) \\ \midrule
\(Prune_{LG}\)  & 22.677           & 0.7889           & 0.2641              & 182.2                                                                & 90.0\%                  & 1.24E-4                \\
\(Prune_{MS}\) & 23.127           & 0.8222           & 0.2232              & 191.3                                                                & 89.5\%                  & 1.21E-4                \\
\(Prune_{RS}\)       & 22.812           & 0.8151           & 0.2395              & 173.2                                                                & 90.5\%                  & \cellcolor{pink!20}1.32E-4                \\
\(Prune_{EG}\)    & 23.154           & 0.8188           & 0.2277              & 164.8                                                                & 91.0\%                  & \cellcolor{pink!50}1.41E-4                \\
\(Prune_{SG}\)    & 23.060           & 0.8165           & 0.2298              & 162.2                                                                & 91.1\%                  & \cellcolor{pink}1.42E-4                \\ \bottomrule
\end{tabular}%
}
\end{table}

\begin{table}[]
\centering
\caption{Result on Deep Blending~\cite{Hedman2018Deep}. The darker the pink color, the better the effectiveness of pruning.}
\label{table:5}
\resizebox{\columnwidth}{!}{%
\begin{tabular}{@{}lcccccc@{}}
\toprule
Prune          & PSNR\(\uparrow\) & SSIM\(\uparrow\) & LPIPS\(\downarrow\) & \begin{tabular}[c]{@{}c@{}}\# Prim.\(\downarrow\)\\ (k)\end{tabular} & Comp. ratio\(\uparrow\) & PSNR/Prim.\(\uparrow\) \\ \midrule
\(Prune_{LG}\)  & 29.070           & 0.8892           & 0.2781              & 421.5                                                                & 85.0\%                  & 6.90E-5                \\
\(Prune_{MS}\) & 29.467           & 0.8987           & 0.2610              & 331.9                                                                & 88.2\%                  & 8.88E-5                \\
\(Prune_{RS}\)       & 29.305           & 0.8990           & 0.2623              & 307.5                                                                & 89.0\%                  & \cellcolor{pink!20}9.53E-5                \\
\(Prune_{EG}\)    & 29.501           & 0.8979           & 0.2627              & 307.2                                                                & 89.0\%                  & \cellcolor{pink!50}9.60E-5                \\
\(Prune_{SG}\)    & 29.420           & 0.8973           & 0.2634              & 305.0                                                                & 89.1\%                  & \cellcolor{pink}9.65E-5                \\ \bottomrule
\end{tabular}%
}
\end{table}

\begin{figure*}
  \centering
  \includegraphics[width=\linewidth]{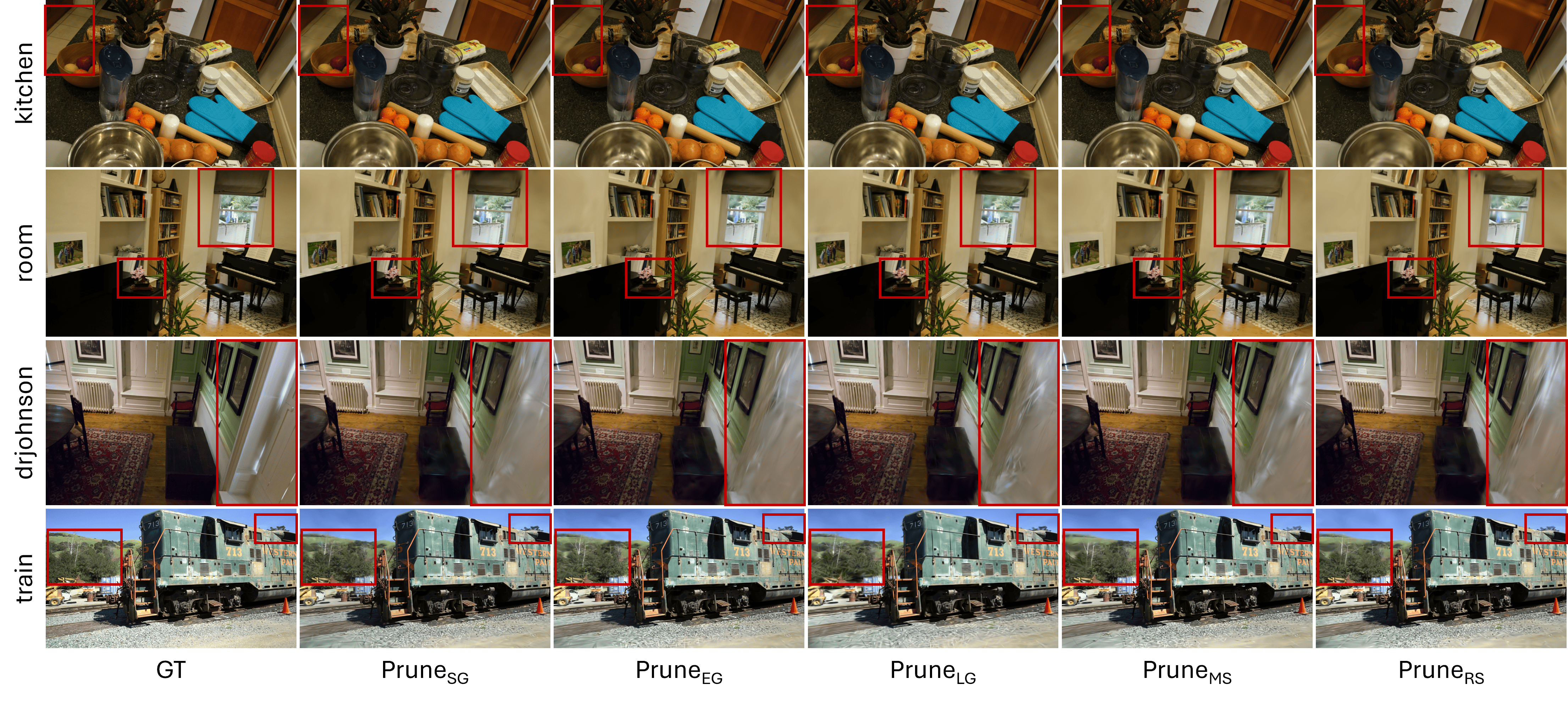}
  \caption{Qualitative result comparison (best viewed in zoom).}
  \label{fig:7}
\end{figure*}

\subsection{Effective pruning technique}
We quantitatively compare the effectiveness of contemporary 3DGS pruning techniques. For pixel-level pruning techniques (i.e., \(Prune_{EG}\)~\cite{Liu2024EfficientGS} and \(Prune_{SG}\)), we select results where the top-k parameter is set to one, compressing the original scene by approximately 90\%. To ensure a fair comparison, we select scene-level pruning results that closely match \(Prune_{SG}\) in the number of primitives from among the swept pruning parameters.

Results on synthetic scenes, as shown in~\cref{table:2}, indicate that \(Prune_{MS}\) outperforms other scene-level pruning techniques. However, the primitive expressiveness metric (PSNR/Prim.) suggests that \(Prune_{SG}\) retains more quality-sensitive primitives, with a notably higher value of 0.8E-4. In comparison to \(Prune_{EG}\), which results in a similar number of primitives, \(Prune_{SG}\) exhibits only a minor PSNR degradation of 0.034 dB. For real scenes from MipNeRF360~\cite{Barron2022Mipnerf360}, pixel-level pruning techniques achieve higher PSNR despite a reduced number of primitives, outperforming scene-level techniques (\cref{table:3}). Notably, \(Prune_{SG}\) demonstrates the highest primitive expressiveness, advancing \(Prune_{EG}\) by 0.06E-5. Across other real scenes, \(Prune_{SG}\) shows minimal quality loss compared to \(Prune_{EG}\) due to more aggressive primitive removal, yet it consistently exhibits the highest expressiveness, as seen in \cref{table:4} and \cref{table:5}.

In \cref{fig:7}, we juxtapose rendered images of each pruning technique. Red boxes highlight areas where they have differences, especially retaining primitives closely matching the ground truth pixel colors. For example, in the kitchen scene, other techniques overly prune primitives representing the floor tiles, whereas \(Prune_{SG}\) retains them. Similarly, in the drjohnson scene, \(Prune_{SG}\) preserves a balanced distribution of primitives across lighted and shaded areas of a door, offering more accurate representation than other techniques.

\section{Conclusion}
In this paper, we categorize Gaussian primitive pruning techniques and demonstrate that pixel-level pruning, coupled with color similarity with blending weight score function (\(Score_{SG\_36}\)), efficiently works for decimating the primitives with minimal performance degradation. From the experiment comparing the techniques, we observe that while scene-level pruning is prone to destructing the scene unless a careful selection of pruning setting is made, pixel-level pruning inherently holds a boundary for pruning setting, preventing excessive removal of Gaussian primitives. Moreover, we notice that the importance functions suggested by prior works rarely consider the impact of primitives' colors. We come up with a variety of score function candidates and test them to find the optimal one. We confirm that the pixel-level pruning technique, using the optimal score function (\(Score_{SG\_36}\)), achieves state-of-the-art efficiency by maximizing primitive expressiveness, reducing training time, and increasing FPS, all with an impressive 90\% compression ratio. We believe our work provides valuable insights for selecting an effective pruning method for future studies.

\paragraph{Limitation}
We mainly focus on pruning effectiveness in this paper. While our study achieves state-of-the-art pruning effectiveness with a 90\% compression ratio and provides a safeguard against catastrophic scene degradation, the rendering quality is slightly worser than that of 3DGS. Moreover, pruning the primitives after the densification stage scarcely helps reduce peak memory usage. We observe that incorporating the proposed pruning technique during the densification stage results in significantly faster training speeds with imperceptible performance drops on the Synthetic NeRF dataset. However, this is not the case for unbounded real-world scenes. Furthermore, like 3DGS, our method does not handle scenes containing various appearances across the training images. Lastly, our method greatly depends on the size of the training data (the total number of pixels of all training images).

\section*{Acknowledgments}
Supported by the Intelligence Advanced Research Projects Activity (IARPA) via Department of Interior/ Interior Business Center (DOI/IBC) contract number 140D0423C0076. The U.S. Government is authorized to reproduce and distribute reprints for Governmental purposes notwithstanding any copyright annotation thereon. Disclaimer: The views and conclusions contained herein are those of the authors and should not be interpreted as necessarily representing the official policies or endorsements, either expressed or implied, of IARPA, DOI/IBC, or the U.S. Government.

{
    \small
    \bibliographystyle{ieeenat_fullname}
    \bibliography{references}
}

\clearpage
\setcounter{page}{1}
\maketitlesupplementary

\section{Score functions}
The rendering equation~\cref{eq:rendering_function} gives good hints for scoring the importance of Gaussian primitives since it directly relates to pixel colors. From the factors and intermediate values of~\cref{eq:rendering_function} (i.e., \(c_i, \bm{\sigma}_i, \alpha_i, \mathcal{G}_i, T_i\)) , we design multiple candidates of effective score function. Because of \(\alpha_i = \bm{\sigma}_i \mathcal{G}_i\), rather than we use directly \(\mathcal{G}_i\), we choose the distance between the center of \(\mathcal{G}_i\) and pixel position. Furthermore, to ensure the factors doesn't dominate in the score calculation, we properly map (squash) their values into \([0, 1]\) by using L1-norm, L2-norm, and cosine similarity function. The squashing functions are:
\begin{align*}
    PC(X_r, X_i) &= \exp(-||X_r - X_i||_2), \\
    CS_{1}(\mathbf{c}_r, c_i) &= (1 - mean(||\mathbf{c}_r - c_i||_1)), \\
    CS_{2}(\mathbf{c}_r, c_i) &= \exp (- mean(||\mathbf{c}_r - c_i||_1)), \\
    CS_{3}(\mathbf{c}_r, c_i) &= (\mathbf{c}_r \cdot c_i) / (||\mathbf{c}_r||_2 * ||c_i||_2),
\end{align*}
where \(X_r\) and \(\mathbf{c}_r\) are, respectively, the pixel coordinates and color of the ray \(r\), \(X_i\) is center of \(\mathcal{G}_i\), \(\cdot\) is vector inner product, and \(*\) is scalar multiplication.

\cref{table:7} lists the full set of functions. Note that the number \(id\) in the function name \(Score_{SG\_{id}}\) is nothing but the score function id.

\begin{table}[]
\centering
\caption{The full list of score functions}
\label{table:7}
\resizebox{\columnwidth}{!}{%
\begin{tabular}{l}
\hline
\(Score_{SG\_1}(\mathbf{G}_i, r) = \bm{\sigma}_i \)  \\
\(Score_{SG\_2}(\mathbf{G}_i, r) = \alpha_i \)  \\
\(Score_{SG\_3}(\mathbf{G}_i, r) = \bm{\sigma}_i * T_i \)  \\
\(Score_{SG\_4}(\mathbf{G}_i, r) = \alpha_i * T_i \)  \\
\(Score_{SG\_5}(\mathbf{G}_i, r) = PC(X_r, X_i) \)  \\
\(Score_{SG\_6}(\mathbf{G}_i, r) = PC(X_r, X_i) * \bm{\sigma}_i \)  \\
\(Score_{SG\_7}(\mathbf{G}_i, r) = PC(X_r, X_i) * \alpha_i \)  \\
\(Score_{SG\_8}(\mathbf{G}_i, r) = PC(X_r, X_i) * \bm{\sigma}_i * T_i \)  \\
\(Score_{SG\_9}(\mathbf{G}_i, r) = PC(X_r, X_i) * \alpha_i * T_i \)  \\
\(Score_{SG\_11}(\mathbf{G}_i, r) = \alpha_i + T_i \) \\
\(Score_{SG\_16}(\mathbf{G}_i, r) = CS_3(\mathbf{G}_i, r) \) \\
\(Score_{SG\_17}(\mathbf{G}_i, r) = CS_3(\mathbf{G}_i, r) * \bm{\sigma}_i \)  \\
\(Score_{SG\_18}(\mathbf{G}_i, r) = CS_3(\mathbf{G}_i, r) * \alpha_i \)  \\
\(Score_{SG\_19}(\mathbf{G}_i, r) = CS_3(\mathbf{G}_i, r) * \bm{\sigma}_i * T_i \)  \\
\(Score_{SG\_20}(\mathbf{G}_i, r) = CS_3(\mathbf{G}_i, r) * \alpha_i * T_i \)  \\
\(Score_{SG\_21}(\mathbf{G}_i, r) = CS_3(\mathbf{G}_i, r) * PC(X_r, X_i) \)  \\
\(Score_{SG\_22}(\mathbf{G}_i, r) = CS_3(\mathbf{G}_i, r) * PC(X_r, X_i) * \bm{\sigma}_i \)  \\
\(Score_{SG\_23}(\mathbf{G}_i, r) = CS_3(\mathbf{G}_i, r) * PC(X_r, X_i) * \alpha_i \)  \\
\(Score_{SG\_24}(\mathbf{G}_i, r) = CS_3(\mathbf{G}_i, r) * PC(X_r, X_i) * \bm{\sigma}_i * T_i \)  \\
\(Score_{SG\_25}(\mathbf{G}_i, r) = CS_3(\mathbf{G}_i, r) * PC(X_r, X_i) * \alpha_i * T_i \)  \\
\(Score_{SG\_32}(\mathbf{G}_i, r) = CS_1(\mathbf{G}_i, r) \) \\
\(Score_{SG\_33}(\mathbf{G}_i, r) = CS_1(\mathbf{G}_i, r) * \bm{\sigma}_i \)  \\
\(Score_{SG\_34}(\mathbf{G}_i, r) = CS_1(\mathbf{G}_i, r) * \alpha_i \)  \\
\(Score_{SG\_35}(\mathbf{G}_i, r) = CS_1(\mathbf{G}_i, r) * \bm{\sigma}_i * T_i \)  \\
\(Score_{SG\_36}(\mathbf{G}_i, r) = CS_1(\mathbf{G}_i, r) * \alpha_i * T_i \)  \\
\(Score_{SG\_37}(\mathbf{G}_i, r) = CS_1(\mathbf{G}_i, r) * PC(X_r, X_i) \)  \\
\(Score_{SG\_38}(\mathbf{G}_i, r) = CS_1(\mathbf{G}_i, r) * PC(X_r, X_i) * \bm{\sigma}_i \)  \\
\(Score_{SG\_39}(\mathbf{G}_i, r) = CS_1(\mathbf{G}_i, r) * PC(X_r, X_i) * \alpha_i \)  \\
\(Score_{SG\_40}(\mathbf{G}_i, r) = CS_1(\mathbf{G}_i, r) * PC(X_r, X_i) * \bm{\sigma}_i * T_i \)  \\
\(Score_{SG\_41}(\mathbf{G}_i, r) = CS_1(\mathbf{G}_i, r) * PC(X_r, X_i) * \alpha_i * T_i \)  \\
\(Score_{SG\_48}(\mathbf{G}_i, r) = CS_2(\mathbf{G}_i, r) \) \\
\(Score_{SG\_49}(\mathbf{G}_i, r) = CS_2(\mathbf{G}_i, r) * \bm{\sigma}_i \)  \\
\(Score_{SG\_50}(\mathbf{G}_i, r) = CS_2(\mathbf{G}_i, r) * \alpha_i \)  \\
\(Score_{SG\_51}(\mathbf{G}_i, r) = CS_2(\mathbf{G}_i, r) * \bm{\sigma}_i * T_i \)  \\
\(Score_{SG\_52}(\mathbf{G}_i, r) = CS_2(\mathbf{G}_i, r) * \alpha_i * T_i \)  \\
\(Score_{SG\_53}(\mathbf{G}_i, r) = CS_2(\mathbf{G}_i, r) * PC(X_r, X_i) \)  \\
\(Score_{SG\_54}(\mathbf{G}_i, r) = CS_2(\mathbf{G}_i, r) * PC(X_r, X_i) * \bm{\sigma}_i \)  \\
\(Score_{SG\_55}(\mathbf{G}_i, r) = CS_2(\mathbf{G}_i, r) * PC(X_r, X_i) * \alpha_i \)  \\
\(Score_{SG\_56}(\mathbf{G}_i, r) = CS_2(\mathbf{G}_i, r) * PC(X_r, X_i) * \bm{\sigma}_i * T_i \)  \\
\(Score_{SG\_57}(\mathbf{G}_i, r) = CS_2(\mathbf{G}_i, r) * PC(X_r, X_i) * \alpha_i * T_i \)  \\
\(Score_{SG\_68}(\mathbf{G}_i, r) = CS_1(\mathbf{G}_i, r) + \alpha_i * T_i \) \\
\(Score_{SG\_75}(\mathbf{G}_i, r) = CS_1(\mathbf{G}_i, r) + \alpha_i + T_i \) \\
\(Score_{SG\_84}(\mathbf{G}_i, r) = CS_2(\mathbf{G}_i, r) + \alpha_i * T_i \) \\
\(Score_{SG\_91}(\mathbf{G}_i, r) = CS_2(\mathbf{G}_i, r) + \alpha_i + T_i \) \\ \hline
\end{tabular}%
}
\end{table}

\section{Preliminary Search of Effective Score Function Candidates}
\begin{figure*}
    \centering
    \includegraphics[width=\linewidth]{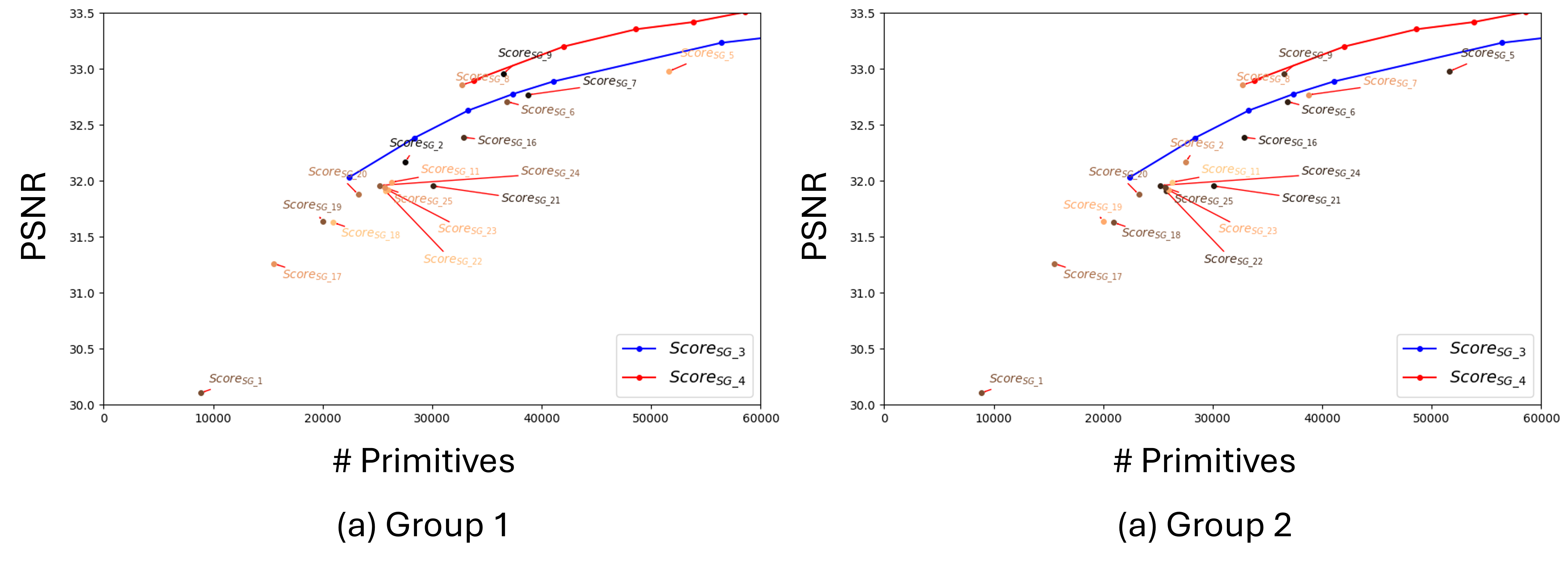}
    \caption{PSNR vs. \#Primitives plot for the top-1 setting of score functions in~\cref{table:7}. The PSNR values are averaged across 8 scenes from Synthetic NeRF~\cite{Mildenhall2020NeRF}. For clarity, the dots are plotted separately in group 1 (a) and group 2 (b).}
    \label{fig:8}
\end{figure*}

\begin{figure*}
    \centering
    \includegraphics[width=\linewidth]{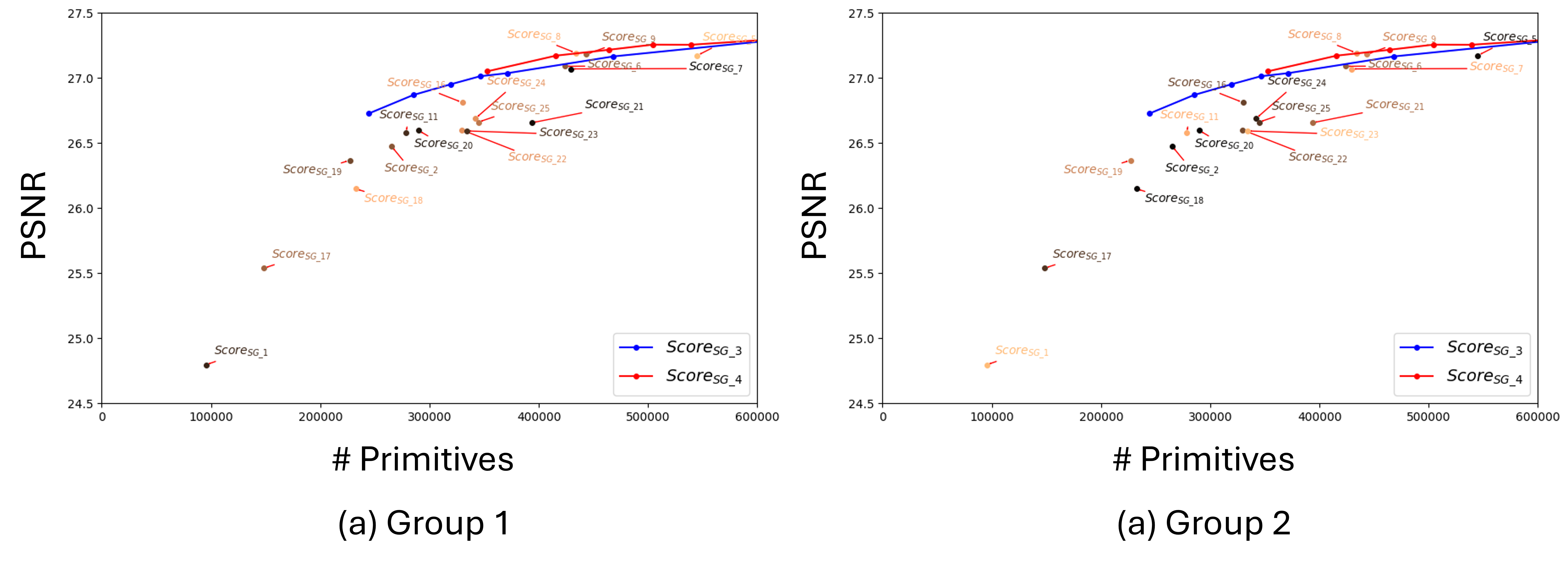}
    \caption{PSNR vs. \#Primitives plot for the top-1 setting of score functions in~\cref{table:7}. The PSNR values are averaged across 9 scenes from MipNeRF360~\cite{Barron2022Mipnerf360}. For clarity, the dots are plotted separately in group 1 (a) and group 2 (b).}
    \label{fig:9}
\end{figure*}

Given the full set of functions from~\cref{table:7}, we performed a pilot study to reduce the number of candidates for the top-k parameter sweep experiment. In the pilot study, we selected two datasets (i.e., Synthetic NeRF~\cite{Mildenhall2020NeRF} and MipNeRF360~\cite{Barron2022Mipnerf360}), as shown in~\cref{fig:8,fig:9}. We trained 3DGS for 30k iterations and performed pruning once at 20k with the top-1 setting for each score function. Other hyperparameters were kept the same as in the vanilla 3DGS. To make the plot legible, we divided the points into two groups. Based on \(Score_{SG\_3}\) and \(Score_{SG\_4}\), we narrowed down the candidates, resulting in~\cref{table:1}.

\begin{figure*}
    \centering
    \includegraphics[width=\linewidth]{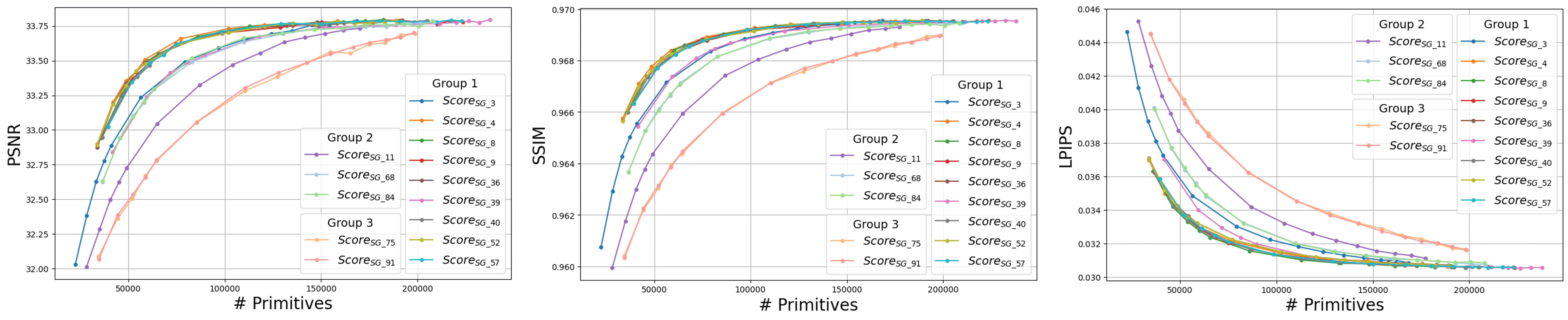}
    \caption{Sweep on the top-k parameter of the selected functions in~\cref{table:1}. The values are averaged over 8 scenes from Synthetic NeRF~\cite{Mildenhall2020NeRF}.}
    \label{fig:10}
\end{figure*}

\begin{figure*}
    \centering
    \includegraphics[width=\linewidth]{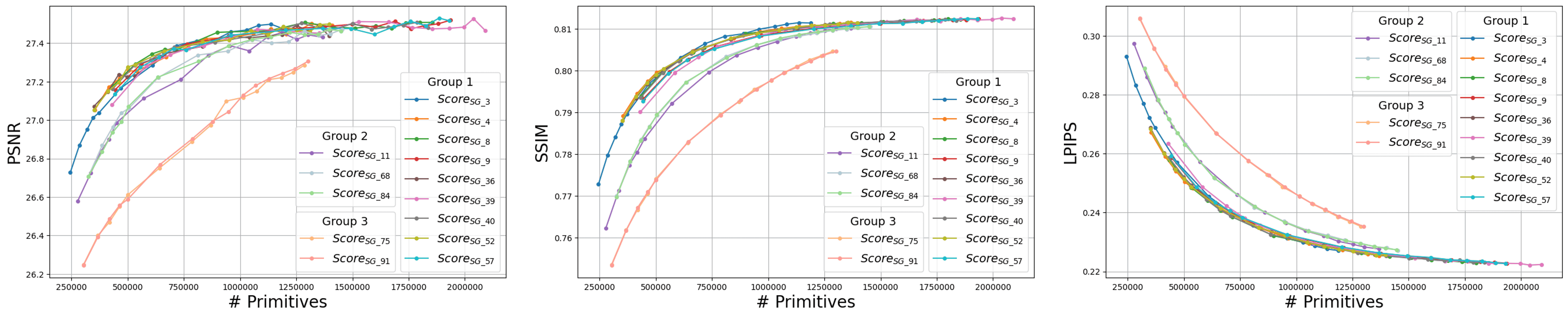}
    \caption{Sweep on the top-k parameter of the selected functions in~\cref{table:1}. The values are averaged over 9 scenes from MipNeRF360~\cite{Barron2022Mipnerf360}.}
    \label{fig:11}
\end{figure*}

\begin{figure*}
    \centering
    \includegraphics[width=\linewidth]{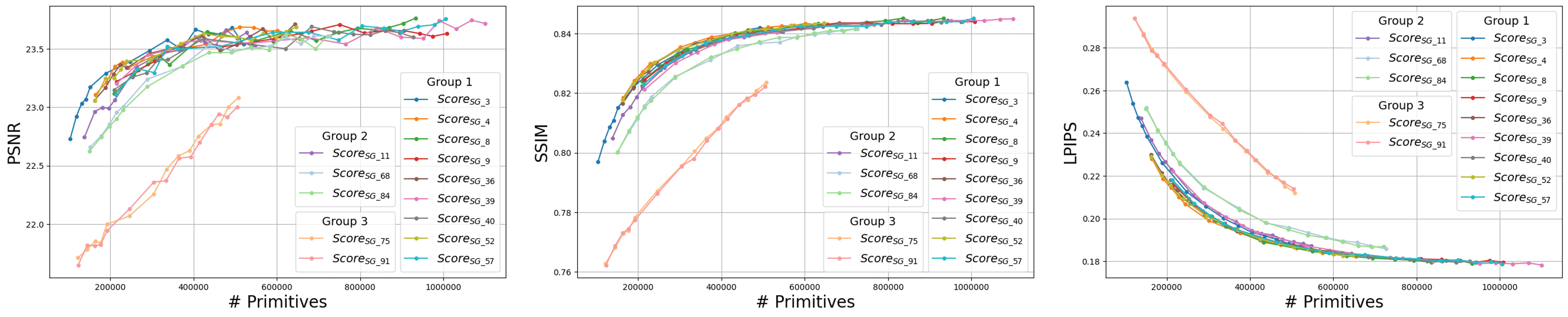}
    \caption{Sweep on the top-k parameter of the selected functions in~\cref{table:1}. The values are averaged over 2 scenes from TanksAndTemples~\cite{Knapitsch2017Tanks}.}
    \label{fig:12}
\end{figure*}

\begin{figure*}
    \centering
    \includegraphics[width=\linewidth]{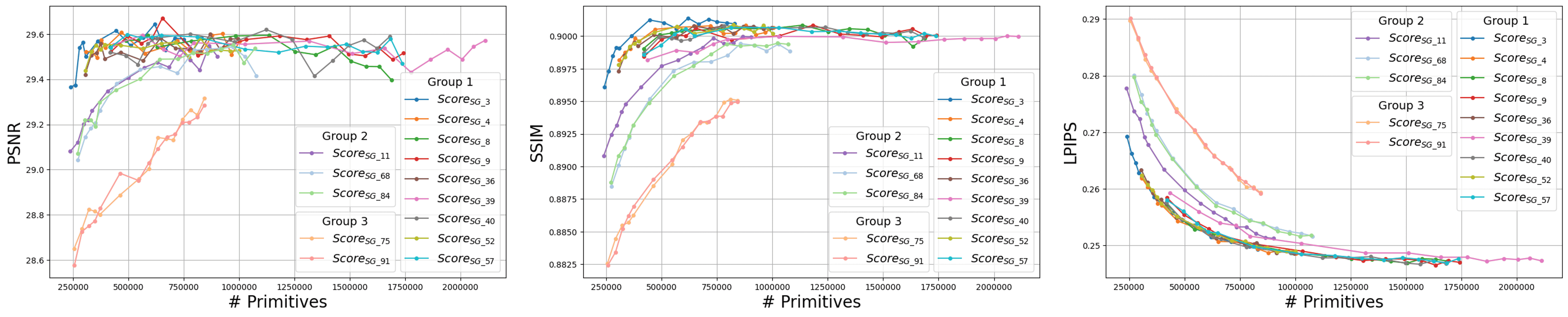}
    \caption{Sweep on the top-k parameter of the selected functions in~\cref{table:1}. The values are averaged over 2 scenes from Deep Blending~\cite{Hedman2018Deep}.}
    \label{fig:13}
\end{figure*}

We further present the top-k parameter sweep results in terms of PSNR, SSIM, and LPIPS metrics across four datasets: Synthetic NeRF~\cite{Mildenhall2020NeRF}, MipNeRF360~\cite{Barron2022Mipnerf360}, TanksAndTemples~\cite{Knapitsch2017Tanks}, and Deep Blending~\cite{Hedman2018Deep}, as shown in~\cref{fig:10,fig:11,fig:12,fig:13}.

\section{Color Similarity Distribution of Primitives}
\begin{figure*}
    \centering
    \includegraphics[height=.95\textheight]{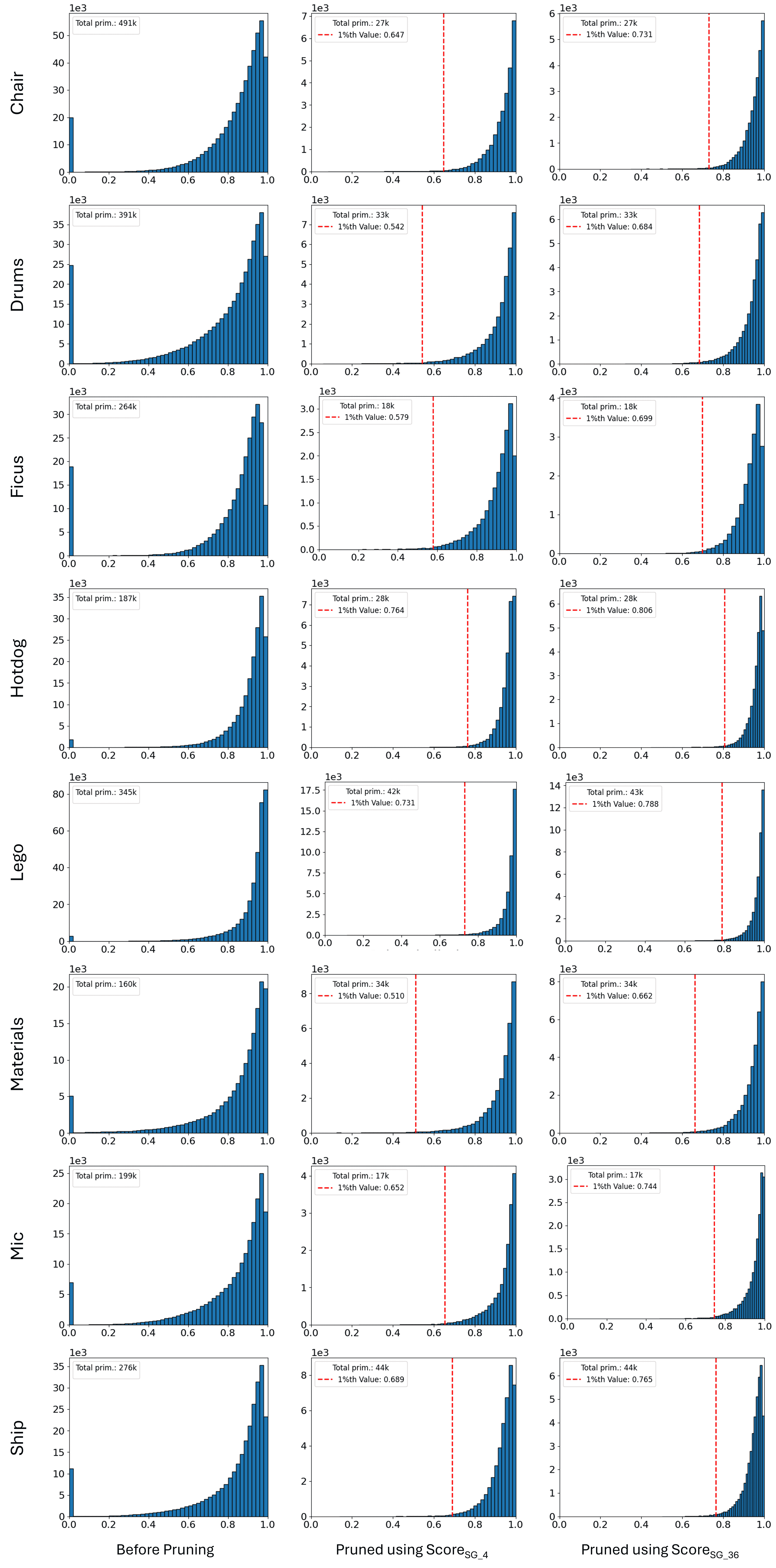}
    \caption{Histogram of primitives' color similarity across 8 scenes from Synthetic NeRF~\cite{Mildenhall2020NeRF}. The X-axis represents color similarity, and the Y-axis shows the number of primitives associated into each bin.}
    \label{fig:14}
\end{figure*}

\begin{figure*}
    \centering
    \includegraphics[width=\linewidth]{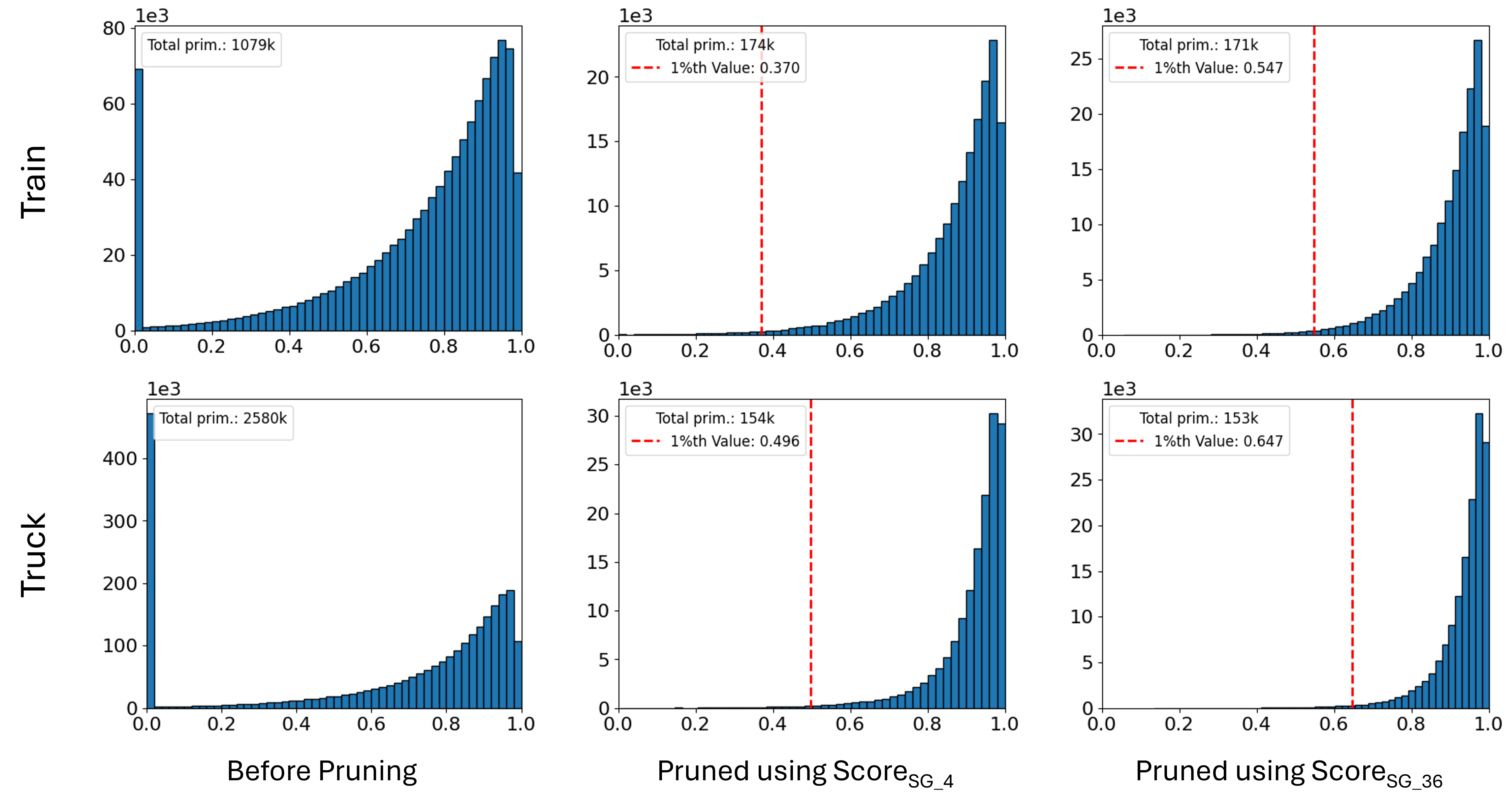}
    \caption{Histogram of primitive's color similarity over 2 scenes from TanksAndTemples~\cite{Knapitsch2017Tanks}. The X-axis represents color similarity, and the Y-axis shows the number of primitives associated into each bin.}
    \label{fig:16}
\end{figure*}

\begin{figure*}
    \centering
    \includegraphics[width=\linewidth]{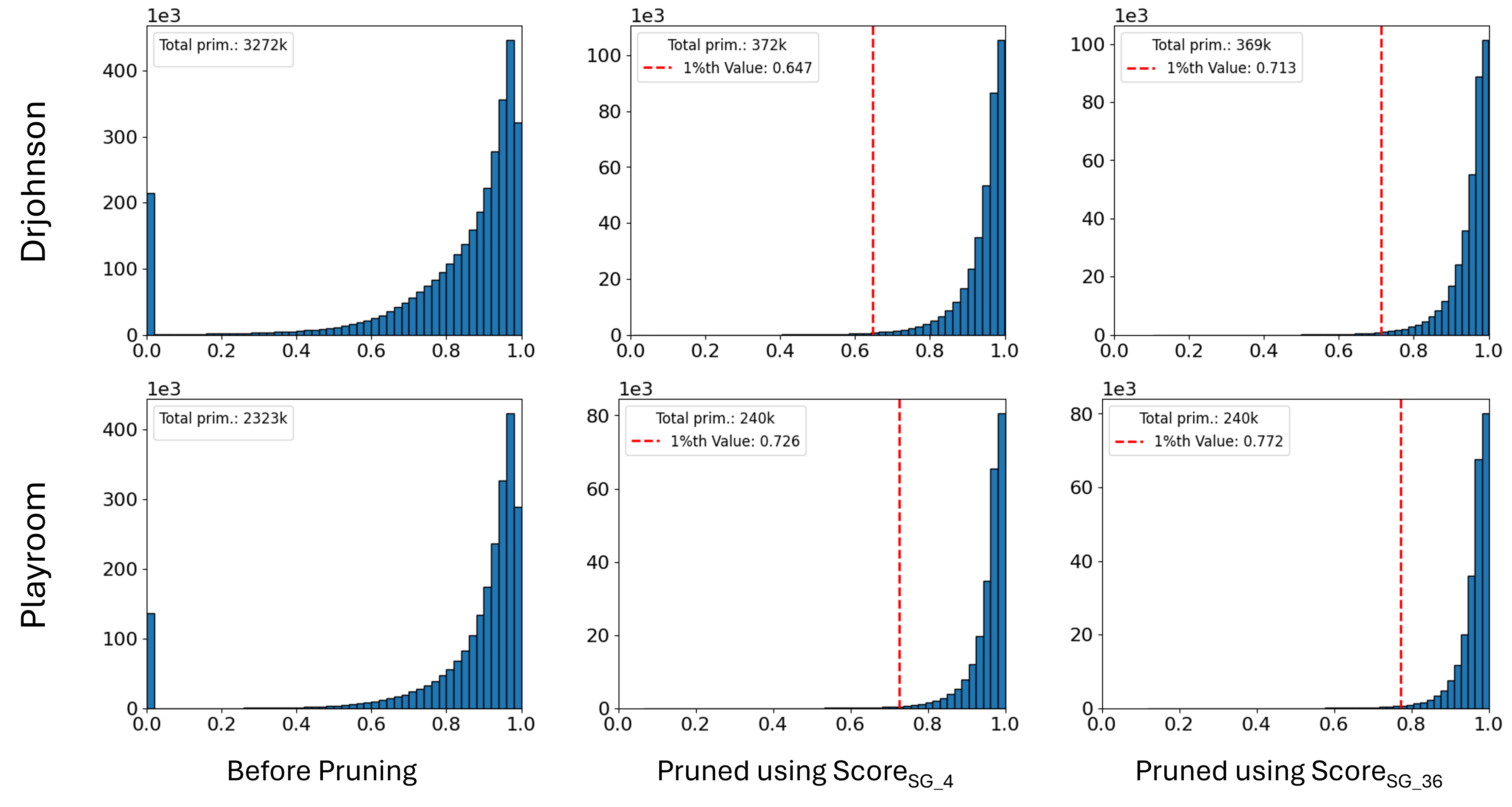}
    \caption{Histogram of primitive's color similarity over 2 scenes from Deep Blending~\cite{Hedman2018Deep}. The X-axis represents color similarity, and the Y-axis shows the number of primitives associated into each bin.}
    \label{fig:17}
\end{figure*}

Across all scenes, we consistently observe that the color similarity distribution is pushed toward 1, as marked by the lower 1\% of primitives' color similarity values through~\cref{fig:14,fig:15,fig:16,fig:17}. Moreover, we observe that the number of primitives remaining after pruning with \(Score_{SG\_36}\) is generally smaller than that with \(Score_{SG\_4}\). This phenomenon implies that \(Score_{SG\_36}\) more effectively selects primitives crucial for maintaining scene rendering quality.

\begin{figure*}
    \centering
    \includegraphics[height=.95\textheight]{figures/fig15.png}
    \caption{Histogram of primitive's color similarity over 9 scenes from MipNeRF360~\cite{Barron2022Mipnerf360}. The X-axis represents color similarity, and the Y-axis shows the number of primitives associated into each bin.}
    \label{fig:15}
\end{figure*}

\section{Rendering Result}
\begin{figure*}
    \centering
    \includegraphics[width=\linewidth]{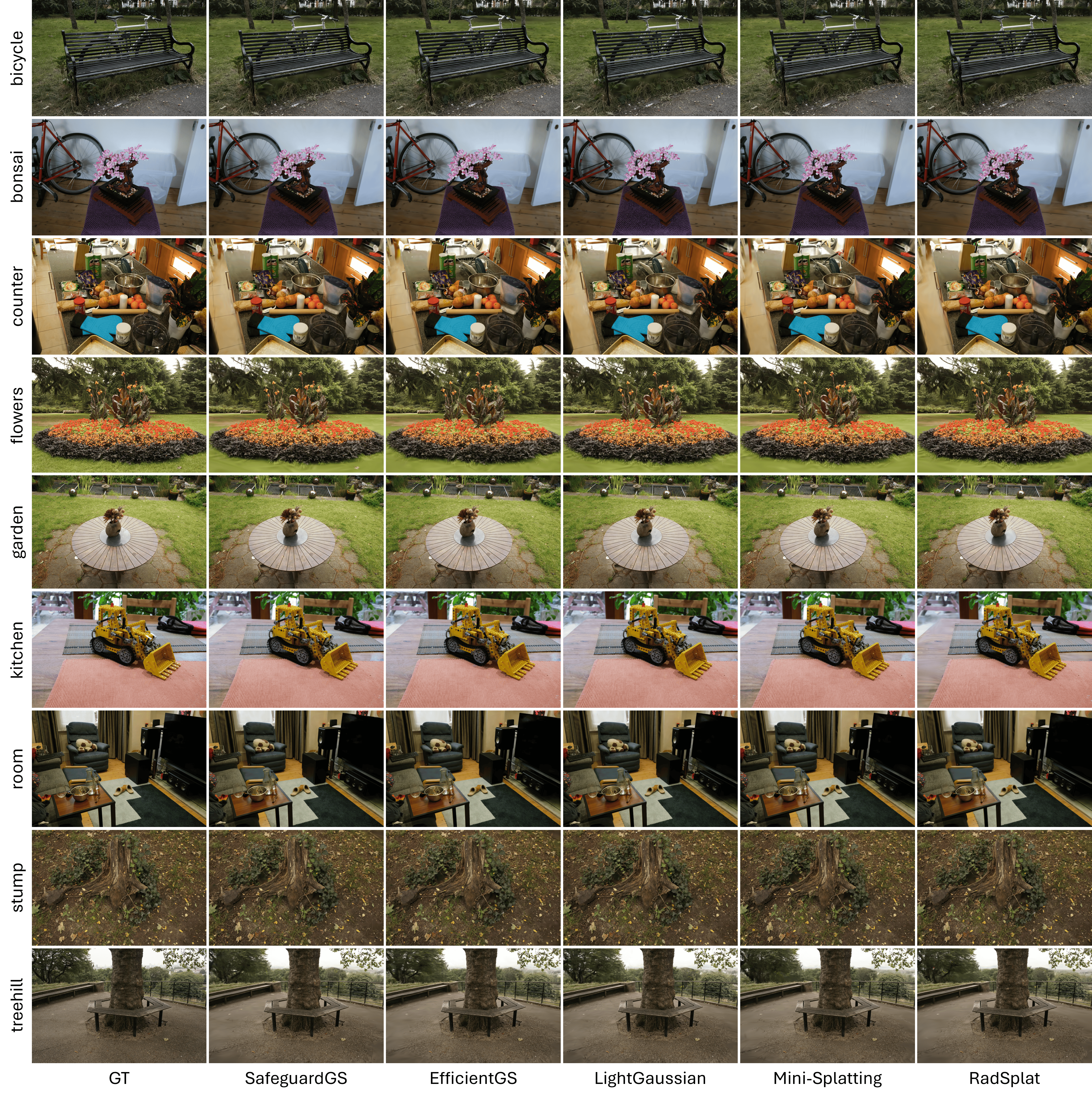}
    \caption{Rendering result of MipNeRF360~\cite{Barron2022Mipnerf360}}
    \label{fig:18}
\end{figure*}

\begin{figure*}
    \centering
    \includegraphics[width=\linewidth]{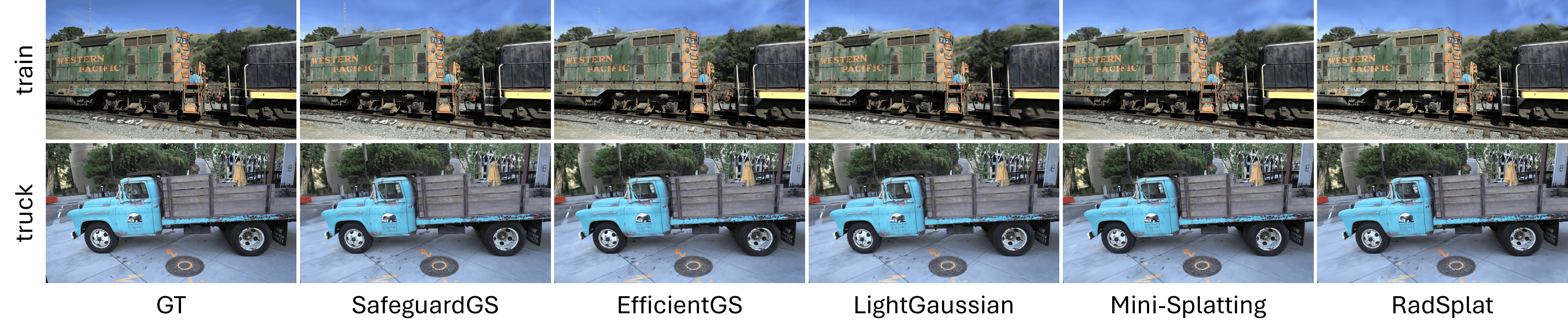}
    \caption{Rendering result of TanksAndTemples~\cite{Knapitsch2017Tanks}}
    \label{fig:19}
\end{figure*}

\begin{figure*}
    \centering
    \includegraphics[width=\linewidth]{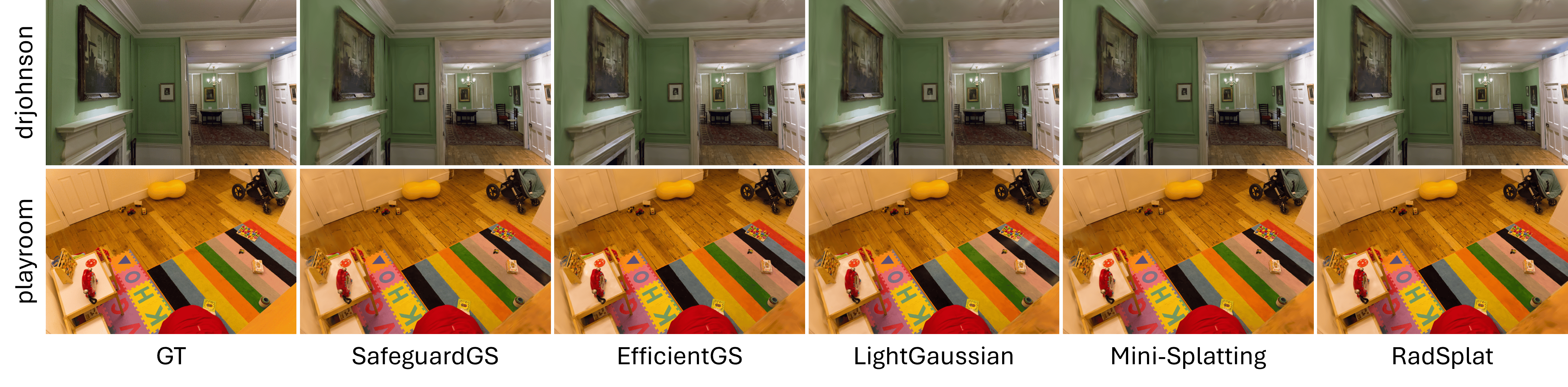}
    \caption{Rendering result of Deep Blending~\cite{Hedman2018Deep}}
    \label{fig:20}
\end{figure*}

Through~\cref{fig:18,fig:19,fig:20}, we present selected rendered images for each scene from MipNeRF360~\cite{Barron2022Mipnerf360}, TanksAndTemples~\cite{Knapitsch2017Tanks}, and Deep Blending~\cite{Hedman2018Deep} to illustrate the visual quality of pruned scenes using SafeguardGS and baseline methods~\cite{Liu2024EfficientGS, Fan2023LightGaussian, Fang2024MiniSplatting, Niemeyer2024RadSplat}.

\end{document}